\pdfoutput=1

\documentclass[11pt]{article}

\usepackage[final]{acl}

\usepackage[T1]{fontenc}
\usepackage[utf8]{inputenc}
\usepackage{times}
\usepackage{latexsym}
\usepackage{microtype}
\usepackage{inconsolata}

\usepackage{amsmath}
\usepackage{amssymb}

\usepackage{graphicx}
\usepackage{float}
\usepackage{booktabs}
\usepackage{multirow}
\usepackage{array}
\usepackage{makecell}
\usepackage{rotating}
\usepackage{pdflscape}
\usepackage{caption}
\usepackage{subcaption}
\usepackage[export]{adjustbox}

\usepackage{xcolor}
\usepackage[table]{xcolor}
\usepackage[most]{tcolorbox}

\usepackage{algorithm}
\usepackage{algorithmicx}
\usepackage{algpseudocode}

\usepackage[percent]{overpic}
\usepackage[normalem]{ulem}
\usepackage{dirtytalk}
\usepackage{enumitem}
\usepackage[]{acronym}
\usepackage{comment}
\usepackage{hyperref}
\usepackage{siunitx}

\definecolor{mypink}{rgb}{0.9686274,0.882352,0.86666}
\definecolor{mygreen}{rgb}{0.819607,0.890196,0.760784}
\definecolor{D55E00}{HTML}{D55E00}
\definecolor{009E73}{HTML}{009E73}
\definecolor{0072B2}{HTML}{0072B2}
\definecolor{Best}{HTML}{DDEBF7}
\definecolor{myboxcolor}{rgb}{0.402,0.402,0.402}

\newtcolorbox{mybox}[1][]{
  enhanced,
  title=#1,
  colback=myboxcolor!3,
  colbacktitle=myboxcolor!3,
  coltitle=black,
  left=4pt,
  right=4pt,
  top=4.5pt,
  bottom=0pt,
  attach boxed title to top left={xshift=8pt, yshift=-7pt},
  boxed title style={frame hidden, size=small, colback=myboxcolor!3},
  sharp corners,
  rounded corners,
  arc=7pt,
}

\title{Min-$k$ Sampling: Decoupling Truncation from Temperature Scaling via \\ Relative Logit Dynamics}

\author{
  Yuanhao Ding$^{1}$~~~
  Meimingwei Li$^{2}$~~~
  Esteban Garces Arias$^{2,3}$~~~
  \textbf{Matthias Aßenmacher}$^{2,3}$\\~~~
  \textbf{Christian Heumann}$^{2}$~~~
  \textbf{Chongsheng Zhang}\thanks{\ \ Corresponding author}$^{1}$\\[1.5ex]
  $^1$School of Computer and Information Engineering, Henan University\\ $^2$Department of Statistics, LMU Munich,
  $^3$Munich Center for Machine Learning (MCML)\\[1.5ex]
  \url{{yhding, cszhang}@henu.edu.cn}, \quad \url{M.Li@campus.lmu.de}\\
  \url{{esteban.garcesarias, matthias, chris}@stat.uni-muenchen.de}
}

\begin{document}

\maketitle

\begin{abstract}

The quality of text generated by large language models depends critically on the decoding sampling strategy. While mainstream methods such as Top-$k$, Top-$p$, and Min-$p$ achieve a balance between diversity and accuracy through probability-space truncation, they share an inherent limitation: extreme sensitivity to the temperature parameter. Recent logit-space approaches like Top-$n\sigma$ achieve temperature invariance but rely on global statistics that are susceptible to long-tail noise, failing to capture fine-grained confidence structures among top candidates. We propose \textbf{Min-$k$ Sampling}, a novel dynamic truncation strategy that analyzes the local shape of the sorted logit distribution to identify "semantic cliffs": sharp transitions from high-confidence core tokens to uncertain long-tail tokens. By computing a position-weighted relative decay rate, Min-$k$ dynamically determines truncation boundaries at each generation step. We formally prove that Min-$k$ achieves strict temperature invariance and empirically demonstrate its low sensitivity to hyperparameter choices. Experiments on multiple reasoning benchmarks, creative writing tasks, and human evaluation show that Min-$k$ consistently improves text quality, maintaining robust performance even under extreme temperature settings where probability-based methods collapse. We make our code, models, and analysis tools publicly available. \footnote{\url{https://github.com/YecanLee/Mink}}
\end{abstract}


\section{Introduction}
\label{sec:introduction}

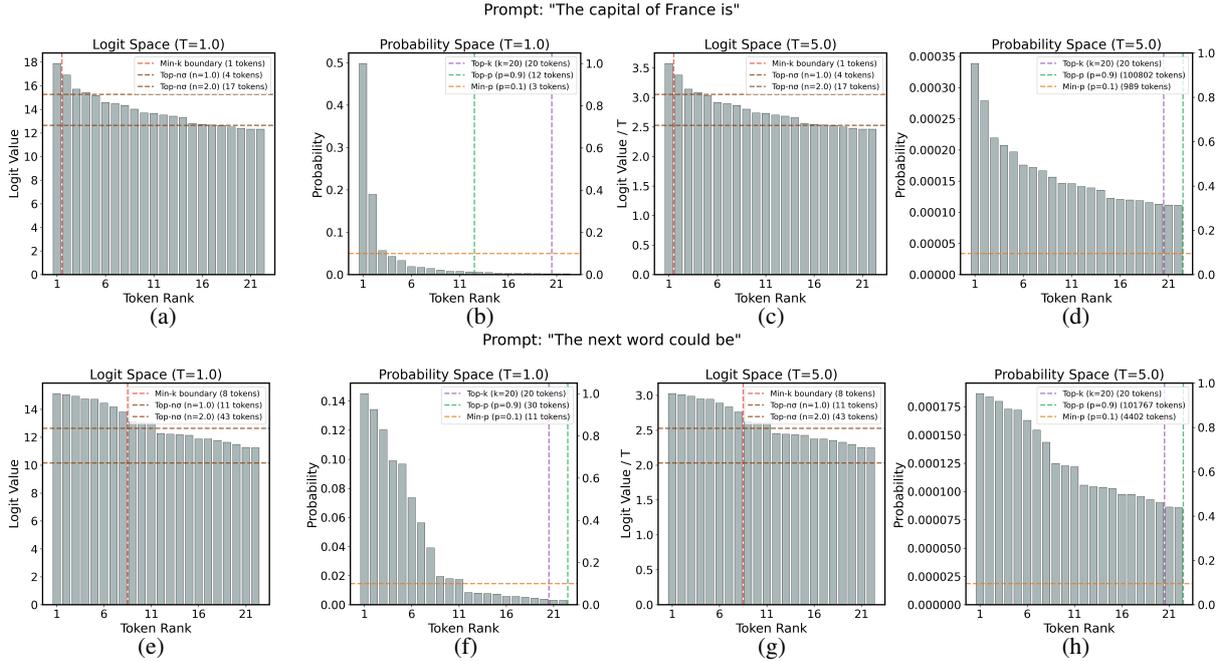
\begin{figure*}[h!]
\centering

\begin{overpic}[width=\linewidth]{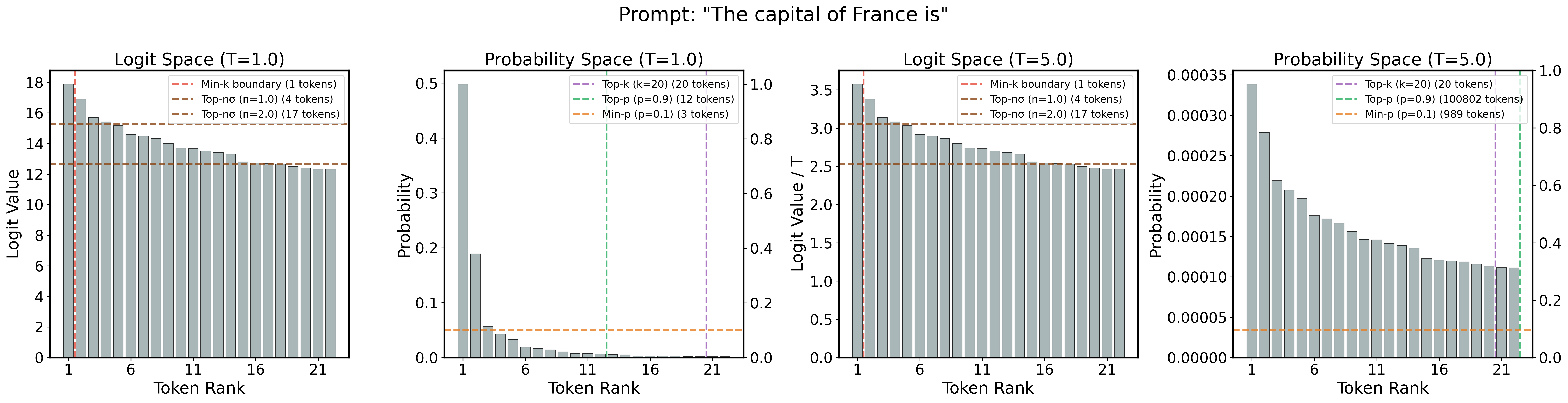}
  \put(12,-1.2){\small (a)}
  \put(38,-1.2){\small (b)}
  \put(62,-1.2){\small (c)}
  \put(87,-1.2){\small (d)}
\end{overpic}

\vspace{0.9em}

\begin{overpic}[width=\linewidth]{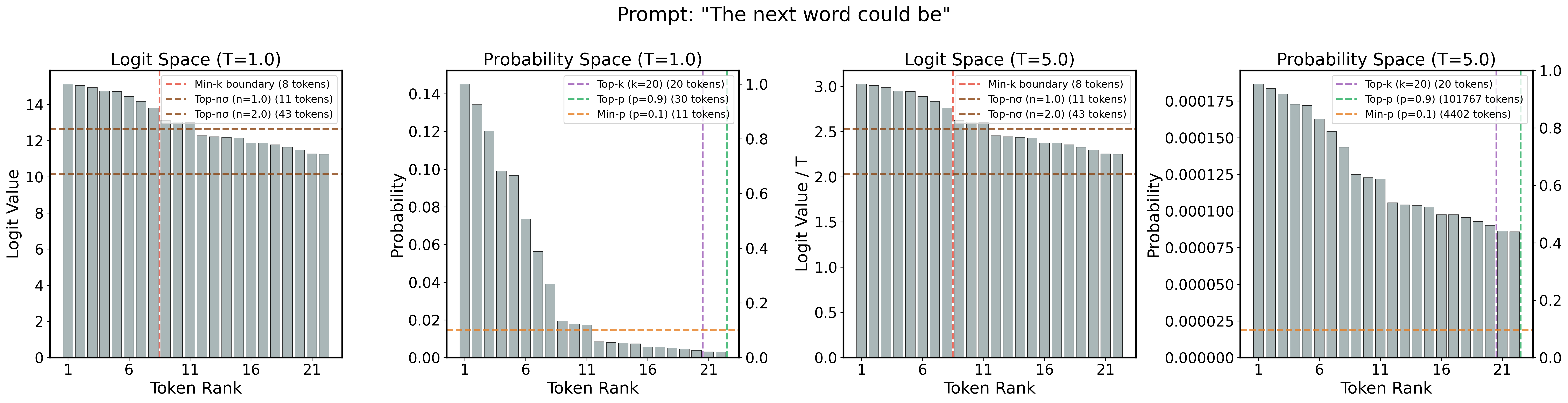}
  \put(11,-1.2){\small (e)}
  \put(37,-1.2){\small (f)}
  \put(62,-1.2){\small (g)}
  \put(87,-1.2){\small (h)}
\end{overpic}

\caption{Comparison of decoding methods under different temperature and confidence conditions. See Appendix~\ref{app:tokendetail} for detailed token information. \textbf{Top:} High-confidence case, where Min-$k$ accurately selects correct tokens and remains temperature-invariant. \textbf{Bottom:} Low-confidence case, where Min-$k$ identifies semantic boundaries while reducing noise tokens.}
\label{fig:one}
\end{figure*}

The quality of texts generated by large language models (LLMs) crucially depends on the sampling strategy used during decoding. Established methods such as Top-$k$ \cite{topk}, Top-$p$ (nucleus sampling) \cite{topp}, and Min-$p$ \cite{minp} strike a balance between diversity and accuracy by truncating the probability space. However, these probability-based approaches share an inherent limitation: they are still extremely sensitive to the \textit{temperature parameter} $T$, whose magnitude controls the probability distribution over candidate tokens and thus directly influences the stochasticity of next-token sampling \cite{ackley1985, chen-ding-2023, bellemare2024}. As illustrated in Figure~\ref{fig:one} (b, d, f, h), using the LLaMA-3-8B-Instruct model \cite{llama3}, we evaluate different probability-based sampling methods under both a high-confidence context \textit{("The capital of France is")} and a low-confidence context \textit{("The next word could be")}. As $T$ increases, the probability distribution becomes flatter, and Top-$p$ as well as Min-$p$ are more likely to (mistakenly) include a large number of low-quality noise tokens within the sampling space. Although Top-$k$ maintains a fixed number of candidates, its reliance on a static $k$ makes it insensitive to dynamic confidence, and its stochasticity remains subject to $T$, which consequently degrades the overall quality of the generated text. As will be shown quantitatively in Section~\ref{sec:noise_analysis}, these methods experience near-total semantic collapse (noise rate > 90\%) when temperature exceeds 2.0.

To alleviate this issue, recent work has introduced the Top-$n\sigma$ sampling \cite{topn}, which operates directly in the logit space. This approach establishes a threshold determined by the maximum logit value and the global standard deviation, thereby achieving invariance to temperature scaling. However, Top-$n\sigma$ relies on the global $\sigma$, making its truncation decision susceptible to interference from background noise within the long-tail logit distribution. Consequently, it fails to accurately locate the "\textit{semantic cliff}"---a phenomenon characterizing the sharp drop in probability mass that separates high-confidence candidates from the unreliable long-tail noise \cite{topp,eta-sampling,zhu2024improving,minp}. Furthermore, the choice of the hyperparameter $n$ directly influences the trade-off between exploration and precision. As illustrated in Figure~\ref{fig:one} (a, c, e, g), under both high-confidence and low-confidence semantic contexts, the default setting of $n=1.0$ (as in the original paper) introduces semantically irrelevant noise, while $n=2.0$ (still within the range of the paper) further enlarges the candidate pool and amplifies the presence of noise tokens. Desired properties of an improved sampling method include \textbf{(1)} preserving temperature invariance and reducing sensitivity to hyperparameter tuning, and \textbf{(2)} faithfully reflecting internal confidence cliffs among top candidates. Existing truncation methods conflate two distinct effects of temperature scaling: diversifying among plausible candidates and admitting noise tokens from the tail. An ideal truncation mechanism should decouple these two effects, allowing temperature to control only diversity within the plausible candidate set.

To this end, we propose \textbf{Min-$k$ Sampling}, a dynamic truncation strategy that directly analyzes the local shape of the sorted logit distribution to determine the optimal truncation point. Our method computes a position-weighted relative decay rate to detect the most pronounced semantic cliff within the logit sequence and applies truncation at this dynamically determined boundary. Because this method is relative in nature, it achieves temperature invariance. More importantly, by focusing on the local head structure of the distribution, it allows the sampling decision to dynamically adapt to the model’s confidence variation at each generation step, effectively identifying semantic boundaries.

\paragraph{Contributions.}

\begin{enumerate}
    \item We propose \textbf{Min-$k$ sampling}, a novel method that analyzes the local shape of the logit distribution to locate the "semantic cliff," enabling adaptive truncation and reducing semantic noise in generated texts.
    
    \item We show that Min-$k$ is \textbf{temperature-invariant} (\S\ref{sec:proof}) and exhibits low sensitivity to hyperparameters, maintaining stable performance across different settings (\S\ref{sec:visual}, \S\ref{subsec:analysis}).
    
    \item Experiments on multiple datasets and the \textbf{Alpaca2.0} benchmark demonstrate that Min-$k$ consistently improves text quality and diversity under both low and high temperatures.
\end{enumerate}


\section{Related Work}
\label{sec:related}
We categorize existing sampling-based decoding strategies into three broad families: probability-based, entropy-based, and logit-space-based:

\paragraph{Probability-based Sampling.}

LLMs generate text by sampling from the probability distribution over their vocabulary. Classical temperature scaling \cite{ackley1985} aims to balance generation diversity and determinism by adjusting the smoothness of this distribution. To further filter long-tail noise, the most straightforward approach, Top-$k$ sampling \cite{topk}, restricts the candidate set to the $k$ most likely tokens, but its fixed size cannot adapt to different contexts. To address this issue, Top-$p$ (nucleus) sampling \cite{topp} dynamically selects the smallest set of tokens whose cumulative probability exceeds a given threshold, thereby achieving context adaptivity. However, Top-$p$ is sensitive to temperature and tends to introduce low-quality tokens at high temperatures. More recently, Min-$p$ \cite{minp} mitigates this issue to some extent via a dynamic threshold, but still fails to fully overcome the challenges posed by high temperatures \cite{topn}. These widely used sampling methods share a common limitation: their decision criteria directly depend on probability values that are themselves temperature-sensitive, leading to unstable performance under extreme temperatures. Moreover, \citet{ma2025} argue that probabilities only capture the relative strength between tokens rather than the model's absolute confidence, which fundamentally limits the reliability of all these methods.

\paragraph{Entropy-based Sampling.}

Entropy-based sampling methods adopt an information-theoretic perspective, utilizing entropy to guide token selection, thereby aiming to achieve a more principled balance between diversity and quality. Mirostat \cite{mirostat} adaptively adjusts the temperature parameter in real time to keep the generation perplexity close to a target value $T$, maintaining consistent generation quality. $\eta$-sampling \cite{eta-sampling} introduces a gradient truncation mechanism based on token-level entropy thresholds, dynamically adjusting the sampling space according to the uncertainty of the distribution. REAL \cite{real} aims to optimize the asymptotic entropy of the sampling process to achieve long-term optimal diversity. In addition, entropy has been incorporated into advanced decoding strategies. Adaptive Contrastive Search \citep[ACS,][]{acs2024} uses local uncertainty (with entropy as a proxy) to dynamically adjust the size of the candidate set and the weight of degeneration penalties. Glocal Uncertainty-Aware Robust Decoding \citep[GUARD,][]{guard2025} combines global and local entropy estimates into a \emph{glocal uncertainty} signal that adaptively tunes contrastive-search parameters, improving both efficiency and text diversity. Guide-to-Generation \citep[G2,][]{g2} employs token-level entropy as a gating signal to selectively apply contrastive logit perturbations from diversity-steering prompts, thereby enhancing variability only when the model exhibits high uncertainty. Locally typical sampling~\citep{typical} takes a complementary approach by selecting tokens whose information content is close to the conditional entropy, aiming to maintain a natural information rate. However, as its selection criterion operates in probability space, it remains inherently temperature-sensitive. Despite their theoretical appeal, these strategies have seen limited adoption in practice. A key reason is that they typically involve more complex implementations, additional computational overhead, and more challenging hyperparameter tuning, while the resulting performance gains often fail to justify the added complexity \cite{zhou-etal-2025-balancing, minp, topn}.

\paragraph{Logit-space-based Sampling.}

Given the inherent temperature sensitivity and reliability issues of probability-space sampling methods, recent work has shifted toward the upstream logit space. The goal is to design strategies that are insensitive to temperature scaling, since the magnitude of logits is believed to more directly reflect the model's accumulated "evidence" and thus encode richer information than probabilities \cite{ma2025}. Top-$n\sigma$ is a representative method in this line of research: it truncates the candidate set by applying a threshold in logit space that depends on the maximum logit and the global standard deviation $\sigma$, thereby achieving temperature invariance \cite{topn}. However, this approach introduces new challenges: Its decision rule heavily relies on $\sigma$, a global statistic that captures the dispersion of logits over the entire vocabulary. Because $\sigma$ is easily dominated by numerous low-score "background noise" tokens that are largely unrelated to the current context, the method may become less precise in capturing subtle confidence differences among top-ranked, high-quality candidates. Moreover, its performance is markedly sensitive to the hyperparameter $n$ (cf. \S\ref{sec:introduction}). Consequently, new methods simultaneously need to preserve temperature invariance, reduce hyperparameter sensitivity, and more faithfully reflect the model's \emph{semantic boundaries} among key candidate tokens.


\section{Methodology}
\label{sec:methodology}

The key idea of \textbf{Min-$k$ Sampling} is to dynamically determine the candidate set size $k$ by analyzing the intrinsic structure of the sorted logit values, thereby identifying the semantic cliff. Unlike conventional truncation strategies, Min-$k$ performs this analysis locally without relying on global statistics such as $\sigma$ or thresholds. Furthermore, the entire design is naturally temperature-invariant, ensuring consistent behavior across temperature settings.

\subsection{Core Principle: Identifying Semantic Cliffs via Weighted Logit Decay}
\label{sec:3.1}

Given the model’s raw logit vector $\mathbf{I} \in \mathbb{R}^{|V|}$ at a decoding step (where $|V|$ is the vocabulary size), our goal is to detect a critical drop within the sorted logits sequence.

\paragraph{Step 1: Sorting and Normalization.}
We first sort the logits in descending order:
\[
\mathbf{I}_{\text{sorted}} = (l_{1}, l_{2}, \dots, l_{|V|}), \quad l_{1} \ge l_{2} \ge \dots \ge l_{|V|}.
\]
To ensure invariance to linear transformations such as temperature scaling, we compute the dynamic range $R_l$\footnote{In our practical implementation, we add a small constant ($\varepsilon = 1\mathrm{e}{-8}$) to the denominator to ensure numerical stability in the rare case of a uniform distribution ($l_1 = l_{|V|}$).}:
\begin{equation}
R_l = l_{1} - l_{|V|} .
\label{eq:range}
\end{equation}

\paragraph{Step 2: Weighted Relative Decay.}
Empirically, the most meaningful probability drops occur near the head of the distribution. To capture this pattern, we define a weighted relative decay measure:
\begin{equation}
w_i = \frac{(l_i - l_{i+1})}{R_l} \cdot \frac{1}{i},
\label{eq:decay}
\end{equation}
where $w_i$ normalizes the local drop by the dynamic range $R_l$ while weighting earlier positions with $1/i$ to emphasize changes near the head. Both components are critical: our ablation study (cf. Appendix~\ref{app:component}) confirms that omitting the weighting term $1/i$ significantly degrades performance by introducing tail noise, while removing the normalization factor $R_l$ leads to model collapse under high-temperature settings. Furthermore, we empirically validate the choice of the linear decay function in Appendix~\ref{app:decay}, showing that $1/i$ offers superior generalization across tasks compared to logarithmic or quadratic alternatives.

\paragraph{Step 3: Cliff Detection.}
We then locate the position of the steepest relative decay, which marks the semantic boundary:
\begin{equation}
k_{\text{cliff}} = i^* = \arg\max_{i \in \{1, \dots, |V|-1\}} w_i.
\label{eq:cliff}
\end{equation}
This position determines the structural boundary between confident and uncertain tokens, and the initial candidate size is thus $k_{\text{cliff}} = i^*$.

\subsection{Enhancing Robustness: Dynamic Fallback Mechanism}

While the above mechanism performs well when the logits distribution exhibits a clear hierarchy, LLMs sometimes produce extremely flat logits under high uncertainty (e.g., ambiguous inputs). In such cases, adjacent logit values are nearly identical, and the sequence $\{w_i\}$ lacks a clear peak, causing $\arg\max w_i$ to collapse to $i^* = 1$, which undesirably yields $k=1$. To avoid this degeneracy, we design a \textit{dynamic fallback mechanism} based on the empirical observation that the logit range $R_l$ is inversely correlated with model uncertainty—flatter distributions imply smaller $R_l$. We thus define a fallback candidate size:
\begin{equation}
k_{\text{fallback}} = \lfloor \tau \cdot R_l^{-1} \rfloor,
\label{eq:fallback}
\end{equation}
where $\tau$ is a small constant hyperparameter activated only when the distribution is nearly uniform (i.e., when the dynamic range $R_l$ is minimal). As shown later in Appendix \ref{sec:visual} and \S\ref{subsec:analysis}, the algorithm is largely insensitive to $\tau$; our ablation study (Appendix~\ref{app:component}) also shows that removing this fallback mechanism does not affect performance on structured reasoning tasks. This empirically confirms that the primary weighted-decay method (\S\ref{sec:3.1}) is sufficiently robust to identify semantic boundaries in most cases. However, we retain the fallback mechanism as a safeguard to prevent collapse (i.e., $k=1$) in high-entropy scenarios without a distinct cliff.

\subsection{Final Candidate Set and Sampling Procedure}

The final candidate size $k$ is determined as the maximum of the cliff-based and fallback mechanisms:
\begin{equation}
k = \max(k_{\text{cliff}}, \, k_{\text{fallback}}).
\label{eq:final_k}
\end{equation}

This ensures that when a distinct semantic cliff exists, $k_{\text{cliff}}$ dominates, while in degenerate cases, the fallback mechanism provides a reasonable minimum level of exploration. After determining $k$, we retain the top-$k$ logits and set all remaining logits to $-\infty$:
\begin{equation}
l_i =
\begin{cases}
l_i, & \text{if } i \le k,\\
-\infty, & \text{otherwise.}
\end{cases}
\end{equation}
The truncated logits vector is then scaled by temperature $T$ and passed through the softmax function to form the final sampling distribution:
\begin{equation}
P(x_i) = \frac{\exp(l_i / T)}{\sum_{j=1}^{|V|}\exp(l_j / T)}.
\end{equation}
Because $k$ is independent of $T$, Min-$k$ maintains strict temperature invariance.

\subsection{Temperature Invariance}
\label{sec:proof}
To guarantee consistent decoding behavior across different temperature settings, we formally prove that Min-$k$ sampling is strictly temperature-invariant. That is, for any temperature coefficient $T>0$, the final candidate set size $k$ and the corresponding token indices remain unchanged.

\noindent\textbf{Proposition (Temperature Invariance).}
Let the original logit vector be $\mathbf{I} \in \mathbb{R}^{|V|}$ and let the temperature-scaled logits be $\mathbf{I}' = \mathbf{I} / T$, where $T>0$ is the temperature coefficient. Let $\mathcal{K}(\mathbf{I}) \subseteq \{1,2,\dots,V\}$ denote the candidate token indices selected by Min-$k$ on the logits $\mathbf{I}$. Then for any $T>0$, we have:
\begin{equation}
\mathcal{K}(\mathbf{I}) = \mathcal{K}(\mathbf{I}').
\label{eq:invariance}
\end{equation}

\noindent\textbf{Proof.}
Let $\mathbf{I}_{\text{sorted}} = (l_1, l_2, \dots, l_{|V|})$
be the logits sorted in descending order with the dynamic range as defined in Eq. \eqref{eq:range} and the normalized relative decay (without weighting, cf. Eq. \eqref{eq:decay}) defined as
\begin{equation}
d_i = \frac{l_i - l_{i+1}}{R_l}, \quad i = 1,2,\dots,|V|-1.
\label{eq:proof_decay}
\end{equation}
The Min-$k$ method uses the \textit{weighted} decay defined in Eq.~\eqref{eq:decay} to identify the steepest "semantic cliff".

Now consider temperature scaling of the logits, which results in $l'_i = l_i/T$ and preserves the ordering (since $T>0$), i.e., $l'_1 \ge l'_2 \ge \dots \ge l'_V$, such that the new dynamic range becomes:
\begin{equation}
R_{l'} = l'_1 - l'_V 
       = \frac{l_1 - l_V}{T} 
       = \frac{R_l}{T}.
\label{eq:scaled_range}
\end{equation}

This results in the normalized relative decay for the scaled logits being equal to that for the raw logits
\begin{align}
d'_i &= \frac{l'_i - l'_{i+1}}{R_{l'}} 
    = \frac{(l_i/T - l_{i+1}/T)}{R_l/T} \\
    &= \frac{l_i - l_{i+1}}{R_l}\notag
    = d_i
    \label{eq:scaled_decay}
\end{align}
and hence, the weighted decay remains unchanged:
\begin{equation}
w'_i = \frac{d'_i}{i} = \frac{d_i}{i} = w_i.
\label{eq:scaled_w}
\end{equation}
Therefore, the position of the maximum $i^*$ is invariant with respect to $T$:
\begin{equation}
i^{*'} = \arg\max_i w'_i = \arg\max_i w_i = i^*.
\end{equation}
Since both the ordering of tokens and the cut-off size $i^*$ are invariant to $T$, the selected candidate set $\mathcal{K}$ is identical, as outlined in Algorithm~\ref{alg:min-k}.

\begin{algorithm}[t]
\caption{Min-$k$ Sampling}
\label{alg:min-k}
\begin{algorithmic}[1]
\State \textbf{Input:} Context $x$, $T$, fallback $\tau$
\State \textbf{Output:} Next token
\State Compute logits: $l \gets \mathrm{LLM}(x)$
\State Sort logits: $l \gets \mathrm{sort}(l)$ 
\State Find rank $i^*$ with maximum relative drop:
\[
  i^* \gets \arg\max_i \frac{\,l_i-l_{i+1}\,}{\,i \cdot (\max(l)-\min(l))\,}
\]
\State Determine candidate size $k$:
\[
  k \gets \mathrm{int}\!\left(\max\!\left(i^*,\, \frac{\tau}{\max(l)-\min(l)}\right)\right)
\]
\State Filter logits: $l_j \gets \begin{cases} l_j & \text{if } j \in \{l_1,\dots,l_k\} \\ -\infty & \text{otherwise} \end{cases}$
\State Scale logits: $l' \gets l/T$
\State \ $p \gets \mathrm{softmax}(l')$
\State \ return next token from distribution $p$
\end{algorithmic}
\end{algorithm}





\section{Experimental Setup}
\label{sec:exp_setup}

\paragraph{Models.}
We evaluate our method on several models from the LLaMA-3 \cite{llama3} family, including LLaMA-3-8B-Instruct, which serves as our primary evaluation model, and LLaMA-3-70B-Instruct. We also include Qwen3 \cite{qwen3} models, specifically Qwen3-4B-Instruct and Qwen3-30B-Instruct, for completeness.

\paragraph{Datasets.}
Following the setup of \citet{topn}, we evaluate Min-$k$ on both reasoning and creative writing datasets. For reasoning tasks, we include AQuA \citep[254 samples,][]{aqua}, GSM8K \citep[1319 samples,][]{gsm8k}, GPQA-main \citep[448 samples,][]{gpqa}, and MATH500 \citep[500 samples,][]{math500,math}. Each dataset is converted into an open-ended generation task for fair comparison (cf. Appendix \ref{a:reasoning}). For creative writing \cite{minp}, we select 500 diverse samples designed to assess creativity, depth, quality, and robustness (cf. Appendix \ref{a:creative}).

\paragraph{Baselines.}
We compare Min-$k$ against several widely used decoding strategies Top-$k$ \cite{topk} ($k=20$), Top-$p$ \cite{topp} ($p=0.9$), Mirostat \cite{mirostat} ($\tau=5.0$), $\eta$-sampling \cite{eta-sampling} ($\eta=9 \times 10^{-4}$), Min-$p$ \cite{minp} ($p=0.1$), and Top-$n\sigma$ \cite{topn} ($n=1.0$). The hyperparameters are chosen as optimal according to previous empirical studies and implementation guidelines \cite{eta-sampling, minp, siml, topn, garces-arias-etal-2025-decoding}. For Min-$k$, we use a default fallback parameter of $\tau=3.0$, which was shown in Section~\ref{sec:proof} and Section~\ref{subsec:analysis} to have a negligible impact (cf. Appendix~\ref{app:component}).

\paragraph{Evaluation Metrics.}
For reasoning tasks, we adopt the \textbf{Exact Match (EM)} metric to assess accuracy. For creative writing tasks, we follow the evaluation framework of Alpaca2.0 \cite{alpacaeval} using DeepSeek-V3.2-Exp \cite{dsv32} to compute win rates against greedy decoding outputs (cf. Appendix \ref{a:creative}).

\paragraph{Human Evaluation}
To assess generation quality, we conducted a human evaluation with four native English speakers on 200 pairs of text continuations. Raters indicated a preference for Method A, Method B, or a tie (detailed guidelines in Appendix~\ref{app:F}). Annotators were compensated at \$20/hour. The presentation order of the outputs was randomized to avoid positional bias. We report inter-rater agreement via Fleiss' Kappa and statistical significance using exact binomial tests.

\begin{table*}[!htb]
\centering
\resizebox{1\textwidth}{!}{\begin{tabular}{clcccccccccccc}
\toprule
\multirow{2}{*}{Dataset} & Model & \multicolumn{6}{c}{LLaMA3-8B-Instruct} & \multicolumn{6}{c}{Qwen3-4B-Instruct}  \\
\cmidrule(lr){3-8} \cmidrule(lr){9-14}
 & $T$ & 1.0 & 2.0 & 3.0 & 4.0 & 5.0 & 10.0 &1.0 & 2.0 & 3.0 & 4.0 & 5.0 & 10.0  \\
\midrule
\multirow{5}{*}{AQuA} 
 & Top-$k$ & 46.85 & 40.94 & 17.72 & 8.66 & 6.69 & 0.00 & 75.20 & 77.56 & 71.65 & 62.20 & 44.09 & 0.00 \\
 & Top-$p$ & 49.61 & 31.10 & 12.99 & 3.54 & 1.18 & 0.00 & 77.95 & 77.17 & 70.87 & 52.76 & 19.29 & 0.00 \\
 & Min-$p$ & \cellcolor{Best}{\textbf{50.00}} & 32.68 & 24.02 & 7.48 & 7.08 & 0.00 & 77.56 & 76.77 & 74.41 & 69.69 & 55.12 & 0.00 \\
 & Top-$n\sigma$ & 48.03 & 47.24 & \cellcolor{Best}{\textbf{46.06}} & 46.06 & 44.49 & \cellcolor{Best}{\textbf{49.61}} & 78.35 & 77.17 & 77.17 & 78.35 & 77.17 & 77.17 \\ 
 & Min-$k$ & \cellcolor{Best}{\textbf{50.00}} & \cellcolor{Best}{\textbf{48.03}} & 44.49 & \cellcolor{Best}{\textbf{52.76}} & \cellcolor{Best}{\textbf{50.39}} & 46.06 & \cellcolor{Best}{\textbf{79.13}} & \cellcolor{Best}{\textbf{79.13}} & \cellcolor{Best}{\textbf{79.13}} & \cellcolor{Best}{\textbf{79.92}} & \cellcolor{Best}{\textbf{78.35}} & \cellcolor{Best}{\textbf{79.13}} \\
 \hline
 \multirow{5}{*}{GPQA-main} 
 & Top-$k$ & 28.35 & 26.79 & 15.18 & 5.58 & 2.23 & 0.00 & 17.19 & 18.75 & 12.28 & 9.60 & 3.57 & 0.00 \\ 
& Top-$p$ & 28.57 & 21.21 & 7.59 & 0.45 & 0.00 & 0.00 & 17.19 & 16.07 & 10.71 & 5.36 & 3.57 & 0.00 \\
& Min-$p$ & \cellcolor{Best}{\textbf{29.02}} & 26.34 & 21.21 & 6.92 & 2.68 & 0.00 & \cellcolor{Best}{\textbf{19.20}} & 16.96 & 14.73 & 10.04 & 5.58 & 0.00 \\
& Top-$n\sigma$ & 26.56 & 29.69 & \cellcolor{Best}{\textbf{29.02}} & 26.79 & 27.46 & 28.35 & 17.41 & 18.75 & \cellcolor{Best}{\textbf{17.86}} & 17.41 & 17.41 & 16.07 \\
& Min-$k$ & \cellcolor{Best}{\textbf{29.02}} & \cellcolor{Best}{\textbf{30.80}} & 26.56 & \cellcolor{Best}{\textbf{27.46}} & \cellcolor{Best}{\textbf{29.24}} & \cellcolor{Best}{\textbf{29.69}} & 18.97 & \cellcolor{Best}{\textbf{18.97}} & 17.41 & \cellcolor{Best}{\textbf{19.20}} & \cellcolor{Best}{\textbf{18.53}} & \cellcolor{Best}{\textbf{18.97}} \\
 \hline
\multirow{5}{*}{GSM8K} 
& Top-$k$ & 75.44 & 54.13 & 9.86 & 1.06 & 0.08 & 0.00 & 93.33 & 92.72 & 92.49 & 89.61 & 79.08 & 0.00 \\ 
& Top-$p$ & 76.27 & 46.63 & 2.50 & 0.15 & 0.00 & 0.00 & 92.80 & \cellcolor{Best}{\textbf{93.56}} & 91.74 & 88.76 & 56.25 & 0.00 \\
& Min-$p$ & 75.36 & 62.70 & 28.81 & 3.94 & 0.38 & 0.00 & 93.18 & 92.80 & 91.74 & 89.92 & 85.14 & 0.45 \\
& Top-$n\sigma$ & 75.44 & 73.69 & 72.55 & 73.09 & 74.07 & 73.77 & 93.40 & 93.03 & \cellcolor{Best}{\textbf{92.87}} & 92.87 & 92.72 & \cellcolor{Best}{\textbf{93.18}} \\
& Min-$k$ & \cellcolor{Best}{\textbf{77.39}} & \cellcolor{Best}{\textbf{76.65}} & \cellcolor{Best}{\textbf{76.02}} & \cellcolor{Best}{\textbf{76.15}} & \cellcolor{Best}{\textbf{76.48}} & \cellcolor{Best}{\textbf{74.79}} & \cellcolor{Best}{\textbf{93.63}} & 93.10 & 92.65 & \cellcolor{Best}{\textbf{93.18}} & \cellcolor{Best}{\textbf{93.56}} & 93.10 \\
 \hline
 \multirow{5}{*}{MATH500} 
 & Top-$k$ & 21.40 & 11.60 & 2.20  & 0.20  & 0.00  & 0.00  & 60.00 & 58.80 & 55.20 & 42.60 & 20.40 & 0.00 \\ 
 & Top-$p$ & 22.00 & 9.20  & 0.60  & 0.00  & 0.00  & 0.00  & \cellcolor{Best}{\textbf{60.60}} & 58.80 & 50.00 & 29.20 & 7.20  & 0.00 \\
 & Min-$p$ & 22.00 & 13.40 & 3.40  & 0.00  & 0.00  & 0.00  & \cellcolor{Best}{\textbf{60.60}} & 58.80 & 55.40 & 48.80 & 33.20 & 0.00 \\
 & Top-$n\sigma$ & \cellcolor{Best}{\textbf{24.60}} & \cellcolor{Best}{\textbf{23.40}} & 21.80 & 20.20 & 17.60 & 21.20 & 60.40 & 57.40 & 57.80 & 57.20 & 57.60 & 57.40 \\
 & Min-$k$ & 24.00 & 23.00 & \cellcolor{Best}{\textbf{25.00}} & \cellcolor{Best}{\textbf{22.00}} & \cellcolor{Best}{\textbf{22.00}} & \cellcolor{Best}{\textbf{21.60}} & 59.20 & \cellcolor{Best}{\textbf{59.20}} & \cellcolor{Best}{\textbf{58.80}} & \cellcolor{Best}{\textbf{59.80}} & \cellcolor{Best}{\textbf{59.40}} & \cellcolor{Best}{\textbf{59.00}} \\

\bottomrule
\end{tabular}}
\caption{Exact Match (\%) for different temperatures (1.0--10.0) and sampling strategies on AQuA, GPQA-main, GSM8K and MATH500 using the LLaMA-3-8B-Instruct and Qwen3-4B-Instruct models. The detailed results for LLaMA-3-70B-Instruct and Qwen3-30B-Instruct are provided in Appendix \ref{app:more model}. Best results in \textbf{bold}.}
\label{tab:table1}
\end{table*}

\section{Results}
\label{sec:results}

\subsection{Mathematical Reasoning}
Table \ref{tab:table1} presents a comprehensive performance comparison of various sampling strategies across four challenging mathematical reasoning benchmarks, utilizing two instruction-tuned models of varying scales (LLaMA-3-8B-Instruct and Qwen3-4B-Instruct) under different temperature settings. Overall, the experimental results allow for several interesting observations: Traditional truncation strategies—namely Top-$k$, Top-$p$, and Min-$p$—demonstrate extreme sensitivity to temperature variations. As $T$ increases, these methods suffer from severe performance degradation, culminating in a near-total collapse (with accuracy approaching 0\%) in the high-entropy setting of $T=10.0$. In stark contrast, both Top-$n\sigma$ and Min-$k$ demonstrate superior resilience; even under extremely high-temperature conditions, they effectively circumvent the logical collapse of model outputs, thereby maintaining the feasibility of the reasoning process. Notably, Min-$k$ exhibits superior performance and stability, achieving higher reasoning accuracy across the majority of experimental configurations. This consistent superiority across diverse models and datasets provides compelling evidence of the effectiveness of Min-$k$ in suppressing long-tail distributional noise, enabling it to preserve the model's core reasoning capabilities to the maximum extent while simultaneously ensuring generation quality.

\begin{mybox}[\textbf{Main Takeaways}]
Empirically, Min-$k$ exhibits three key benefits: (1) superior stability in mid-temperature regimes where accuracy and coherence are both crucial, (2) resilience against high-temperature degradation, and (3) improved logical coherence in reasoning-centric tasks. 
\end{mybox}

\begin{table}[!htb]
\centering
\resizebox{0.5\textwidth}{!}{\begin{tabular}{cccccc}
\toprule
\multirow{2}{*}{Method} & \multicolumn{2}{c}{LLaMA3-8B-Instruct} & \multicolumn{2}{c}{Qwen3-4B-Instruct} \\ 
\cmidrule(lr){2-3} \cmidrule(lr){4-5}
 & $T = 1.0$ & $T = 3.0$ & $T = 1.0$ & $T = 3.0$ \\ 
\midrule
Top-$k$ & 50.80 & 1.40 & 50.00 & 13.80 \\
Top-$p$ & 49.60 & 0.00 & 54.60 & 0.40 \\
Mirostat & 48.20 & 1.40 & 48.40 & 18.00 \\
$\eta$-sampling & 52.60 & 0.00 & 49.80 & 0.20 \\
Min-$p$ & 53.40 & 51.60 & 49.40 & 46.60 \\
Top-$n\sigma$ & 52.40 & 51.40 & 54.80 & 50.80 \\ 
Min-$k$ & \cellcolor{Best}{\textbf{58.60}} & \cellcolor{Best}{\textbf{53.60}} & \cellcolor{Best}{\textbf{56.00}} & \cellcolor{Best}{\textbf{55.40}} \\
\hline
 & $T = 5.0$ & $T = 10.0$ & $T = 5.0$ & $T = 10.0$ \\ 
 \cmidrule(lr){2-3} \cmidrule(lr){4-5}
Top-n$\sigma$ & \cellcolor{Best}{\textbf{54.20}} & 50.00 & 50.60 & 51.20 \\ 
Min-$k$ & 53.20 & \cellcolor{Best}{\textbf{52.80}} & \cellcolor{Best}{\textbf{52.20}} & \cellcolor{Best}{\textbf{52.80}} \\
\bottomrule
\end{tabular}
}
\caption{Win rates (\%) on creative writing for both models. Best results in \textbf{bold}.}
\label{tab:creativewriting}
\end{table}

\begin{table}[ht!]
\centering
\resizebox{0.5\textwidth}{!}{\begin{tabular}{lccc}
\toprule
Method & LLaMA3-8B-Instruct & Qwen3-4B-Instruct & Total (\%) \\
\midrule
Top-$n\sigma$ & 33 & \cellcolor{Best}{\textbf{34}} & 67 (33.5)\\
Tie                 & 26 & 32 & 58 (29.0)\\
Min-$k$      & \cellcolor{Best}{\textbf{41}} & \cellcolor{Best}{\textbf{34}} & \cellcolor{Best}{\textbf{75 (37.5)}}\\
\midrule
\textbf{Total}      & 100 & 100 & 200 (100)\\
\bottomrule
\end{tabular}}
\caption{Human evaluation outcomes for Min-$k$ and Top-$n\sigma$, stratified by model. Best results in \textbf{bold}.}
\label{tab:human_eval_by_model}
\end{table}

\subsection{Creative Writing}
To evaluate the generalization capabilities and quality of sampling strategies in open-domain instruction following tasks, Table \ref{tab:creativewriting} reports results from an LLM-as-a-Judge evaluation, with DeepSeek-V3.2-Exp serving as the judge. Following \citet{topn}, we assessed the Win Rate of responses generated by various sampling strategies against a Greedy Decoding baseline on the AlpacaEval dataset. The overall results exhibit a notable performance divergence: As $T$ increases from $1.0$ to $3.0$, the win rates of most methods (Top-$k$, Top-$p$, and Mirostat) plummet to single digits or near-zero levels. This collapse highlights the limitations of traditional static or cumulative probability-based truncation mechanisms in effectively isolating long-tail noise within high-variance distributions. Again, Min-$k$ demonstrates resilience as it not only achieves peak win rates in low-temperature settings ($T=1.0$)—reaching 58.60\% on LLaMA-3 by precisely capturing high-quality candidates—but also establishes a distinct advantage over the Top-$n\sigma$ baseline under extremely high-temperature conditions ($T=10.0$), maintaining a remarkable win rate of 52.80\%.

\begin{mybox}[\textbf{Main Takeaways}]
These results demonstrate the ability of Min-$k$ to maintain high quality, i.e., striking a balance between generative creativity and logical coherence, thereby providing a robust strategy that can adapt to the fluctuating uncertainty inherent in token generation.
\end{mybox}

\subsection{Human Evaluation}
\label{sec:human_eval_alignment}

The human evaluation comprises 200 pairwise comparisons between Min-$k$ and Top-$n\sigma$. Overall, Min-$k$ is preferred in 75 cases (37.5\%), Top-$n\sigma$ in 67 cases (33.5\%), while there are 58 (29.0\%) ties (cf. Table \ref{tab:human_eval_by_model}). This preference for Min-$k$ is more pronounced for LLaMA-3 outputs (41\% vs.\ 33\%) than for Qwen3, where judgments show no clear preference. Excluding ties yields higher agreement (33\% to 75\%) and slightly improved Fleiss' $\kappa$ values ($-0.20$ to $0.33$), though agreement remains in the slight-to-fair range. A binomial test (at $\alpha = 0.05$) revealed no statistically significant difference in human preferences between the two methods.

\subsection{Sensitivity Analysis with Respect to $\tau$}
\label{subsec:analysis}

\begin{figure}[!ht]
\centering
\includegraphics[width=\linewidth]{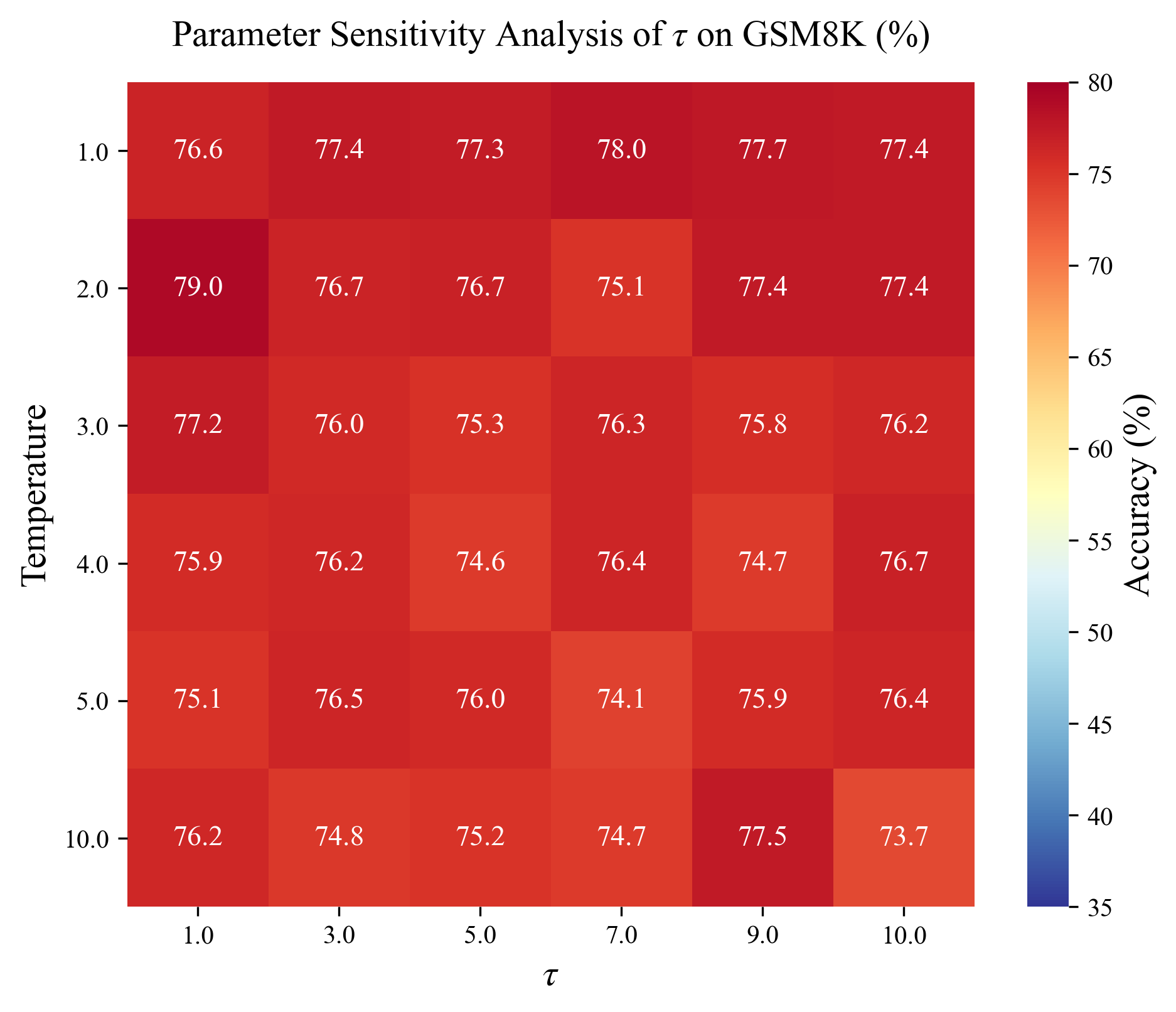}
\caption{Sensitivity analysis of Min-$k$ with respect to $\tau$ on GSM8K. The heatmap shows accuracy (\%) under different combinations of $\tau$ and temperature. The method maintains highly stable performance ($\sim$74-79\%) across the entire tested parameter space, demonstrating strong robustness without a clear degradation threshold even at very high temperatures.}
\label{fig:heatmap}
\end{figure}

To assess the sensitivity of Min-$k$ Sampling with respect to its core hyperparameter $\tau$ and to investigate its robustness under different temperature settings, we conduct a comprehensive grid-search experiment. Using the LLaMA-3-8B-Instruct model, we evaluate combinations of $\tau$ and temperature $T$ (both 1.0 to 10.0) on GSM8K. The results are visualized as a heatmap in Figure~\ref{fig:heatmap}. Our analysis reveals two key insights: Min-$k$ sampling exhibits remarkable overall robustness. As shown in Figure~\ref{fig:heatmap}, across the entire hyperparameter space, the model’s accuracy consistently lies within a very narrow high-performance range (approximately 74\%--79\%). We do not observe any sharp drops in accuracy, indicating that the method is largely insensitive to hyperparameter choices and thus highly convenient for practical deployment. In contrast, the Top-\(n\sigma\) method \cite{topn}, as shown in its original paper’s heatmap, exhibits significantly larger performance fluctuations on GSM8K as the parameter \(n\) varies within its reasonable range.

\begin{mybox}[\textbf{Main Takeaways}]
The ability of Min-$k$ to maintain high consistency across a broad hyperparameter space shows its strong intrinsic stability. This suggests that Min-$k$ is a broadly applicable decoding strategy that significantly reduces reliance on hyperparameter search, while offering improved ease of deployment and generalization relative to the baselines.
\end{mybox}

\subsection{Semantic Noise Analysis}
\label{sec:noise_analysis}

\begin{figure}[!ht]
    \centering
    \includegraphics[width=\linewidth]{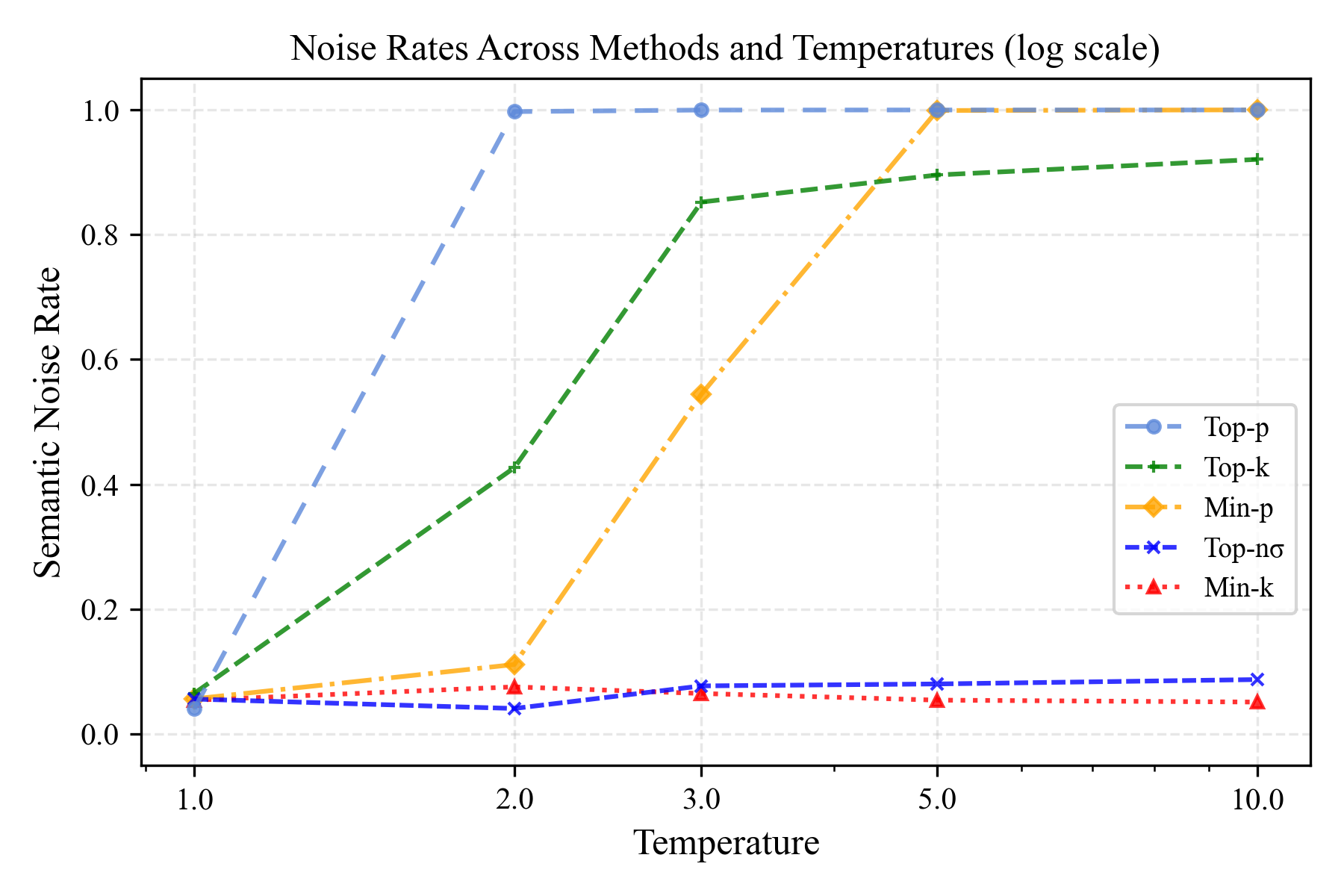} 
    \caption{Semantic Noise Rate Comparison (Log Scale). We evaluate the ratio of incoherent tokens generated by different methods across temperatures ranging from 1.0 to 10.0 on GSM8K (128 samples). While traditional methods collapse (noise rate $\to$ 1.0) as temperature rises, Min-$k$ remains remarkably stable, consistently filtering out long-tail noise without manual tuning.}
    \label{fig:noise_analysis}
\end{figure}

To further investigate why Min-$k$ maintains robustness under high-temperature settings, we conduct a quantitative analysis of the semantic noise rate across different decoding strategies. Unlike traditional metrics that rely on lexical overlap, we define semantic noise as the proportion of generated content that becomes incoherent, repetitive, or irrelevant after a semantic collapse threshold is reached. We employed DeepSeek-V3.2-Exp \cite{deepseekv32} as an impartial judge to identify the exact onset of semantic collapse for each generated response (see Appendix~\ref{app:noise_eval} for detailed evaluation protocols). Figure~\ref{fig:noise_analysis} illustrates the noise rate trends as $T$ increases from 1.0 to 10.0. Standard probability-based methods (Top-$k$, Top-$p$, and Min-$p$) exhibit a sharp increase in noise accumulation as the distribution flattens, with Top-$p$ and Min-$p$ reaching near 100\% noise saturation at $T \ge 2.0$. While Top-$n\sigma$ and Min-$k$ both effectively suppress noise, Min-$k$ even more consistently maintains a negligible noise rate ($< 10\%$) even at extreme temperatures ($T=10.0$), demonstrating its unique ability to dynamically identify and truncate at the true semantic boundary, regardless of distributional smoothness.

\subsection{Computational Efficiency}

To quantify the runtime overhead of Min-$k$, we benchmark all methods on LLaMA-3-8B-Instruct. Table~\ref{tab:latency} reports throughput and per-token latency. Min-$k$ introduces a 3.73\% relative slowdown over greedy decoding, comparable to Top-$p$ (3.46\%) and Top-$n\sigma$ (2.81\%). Although Min-$k$ requires sorting the full logit vector ($\mathcal{O}(|V| \log |V|)$), this operation is highly optimized on modern GPUs and negligible relative to the model's forward pass ($<$0.3\,ms per step on a 32K vocabulary). These results confirm that Min-$k$ is suitable for real-time deployment without specialized kernels.

\begin{table}[t]
\centering
\small
\begin{tabular}{lccc}
\toprule
Method & Tokens/s & Latency (ms) & Slowdown \\
\midrule
Greedy & 39.33 & 25.43 & 0.00\% \\
Top-$k$ & 38.28 & 26.13 & 2.67\% \\
Top-$n\sigma$ & 38.22 & 26.16 & 2.81\% \\
Min-$p$ & 38.10 & 26.24 & 3.11\% \\
Top-$p$ & 37.97 & 26.34 & 3.46\% \\
\rowcolor{Best} Min-$k$ & 37.86 & 26.41 & 3.73\% \\
\bottomrule
\end{tabular}
\caption{Latency benchmark on LLaMA-3-8B-Instruct (single A100). Relative slowdown is computed against greedy decoding.}
\label{tab:latency}
\end{table}


\section{Conclusion}

In this work, we introduced \textbf{Min-$k$ Sampling}, a novel and robust decoding algorithm designed to overcome the inherent temperature sensitivity of probability-based truncation methods for LLM decoding. By analyzing the intrinsic local geometric structure of the sorted logits through a position-weighted decay mechanism, Min-$k$ dynamically locates the semantic boundaries of candidate tokens independently of temperature scaling. Our theoretical analysis confirms its strict temperature invariance, while extensive empirical results across mathematical reasoning and creative writing benchmarks demonstrate its superiority in striking a balance between precision and exploration. Crucially, Min-$k$ serves as a reliable "safety rail" for high-temperature settings, effectively suppressing long-tail semantic noise where traditional methods fail. Min-$k$ provides a new perspective on logit-space truncation, enabling more stable and diverse decoding strategies in large language models.



\section*{Limitations}

While Min-$k$ Sampling demonstrates strong robustness and performance across diverse tasks, we acknowledge two primary limitations:

\noindent (1) Similar to Top-$p$ and Min-$p$, our method relies on sorting or scanning the full logit distribution to compute relative variations, which incurs a marginal computational overhead compared to static Top-$k$ selection, although this is negligible relative to the model's forward pass. 

\noindent (2) The core assumption of Min-$k$---that a semantic boundary manifests as a distinct "cliff" in the logit curve---may be less effective in scenarios requiring the retrieval of extremely long-tail knowledge entities where the correct token possesses a very low probability indistinguishable from noise. In such rare cases, aggressive truncation (by any method) might inadvertently filter out the correct answer. Future work could explore integrating retrieval-augmented mechanisms to mitigate this trade-off.

\section*{Ethics Statement}

We affirm that our research adheres to the \href{https://www.aclweb.org/portal/content/acl-code-ethics}{ACL Ethics Policy}. This work involves the use of publicly available datasets and does not include any personally identifiable information. For our human evaluation, we employed third-party evaluators, ensuring compensation of over \$20 per hour. An ethical concern worth mentioning is the use of language models for text generation, which may produce harmful content, either through intentional misuse by users or unintentionally due to the training data or algorithms. We declare that there are no conflicts of interest that could potentially influence the outcomes, interpretations, or conclusions of this research. All funding sources supporting this study are acknowledged in the acknowledgments section. We have diligently documented our methodology, experiments, and results, and we commit to sharing our code, data, and other relevant resources to enhance reproducibility and further advancements in the field.

\section*{Acknowledgments}

This work was partially supported by the MOE Liberal Arts and Social Sciences Foundation (No.23YJAZH210), Major Program of National Social Science Foundation (No.23\&ZD309), Henan Provincial Center for Outstanding Overseas Scientists (No.GZS2025004), High Level Talent International Training Program of Henan Province (No.GCC2025010), and the Chinese Scholarship Council (Grant No.202308410339). Moreover, Matthias Aßenmacher received funding from the BERD@NFDI consortium in the context of the work of the National Research Data Infrastructure (NFDI) Association. NFDI is funded by the Federal Republic of Germany and the 16 federal states. The BERD@NFDI consortium is supported within NFDI by the German Research Foundation (DFG) – NFDI 27/1-2026, project number 460037581.\\Esteban Garces Arias sincerely thanks the Mentoring Program of the Faculty of Mathematics, Statistics and Informatics at LMU Munich, and the Munich Center for Machine Learning (MCML) for their support. Last but not least, we thank all the anonymous (meta) reviewers for their constructive feedback throughout the review process. 


\bibliography{custom}

@inproceedings{topk,
    title = "Hierarchical Neural Story Generation",
    author = "Fan, Angela  and
      Lewis, Mike  and
      Dauphin, Yann",
    editor = "Gurevych, Iryna  and
      Miyao, Yusuke",
    booktitle = "Proceedings of the 56th Annual Meeting of the Association for Computational Linguistics (Volume 1: Long Papers)",
    month = jul,
    year = "2018",
    address = "Melbourne, Australia",
    publisher = "Association for Computational Linguistics",
    url = "https://aclanthology.org/P18-1082/",
    doi = "10.18653/v1/P18-1082",
    pages = "889--898",
}

@inproceedings{topp,
title={The Curious Case of Neural Text Degeneration},
author={Ari Holtzman and Jan Buys and Li Du and Maxwell Forbes and Yejin Choi},
booktitle={International Conference on Learning Representations(ICLR 2020)},
year={2020},
}

@inproceedings{minp,
  title={Turning up the heat: Min-p sampling for creative and coherent llm outputs},
  author={Nguyen, Minh Nhat and Baker, Andrew and Neo, Clement and Roush, Allen and Kirsch, Andreas and Shwartz-Ziv, Ravid},
  booktitle={International Conference on Learning Representations(ICLR 2025)},
  year={2025}
}

@inproceedings{topn,
    title = "Top-$n\sigma$: Eliminating Noise in Logit Space for Robust Token Sampling of {LLM}",
    author = "Tang, Chenxia  and
      Liu, Jianchun  and
      Xu, Hongli  and
      Huang, Liusheng",
    editor = "Che, Wanxiang  and
      Nabende, Joyce  and
      Shutova, Ekaterina  and
      Pilehvar, Mohammad Taher",
    booktitle = "Proceedings of the 63rd Annual Meeting of the Association for Computational Linguistics (Volume 1: Long Papers)",
    month = jul,
    year = "2025",
    address = "Vienna, Austria",
    publisher = "Association for Computational Linguistics",
    url = "https://aclanthology.org/2025.acl-long.528/",
    doi = "10.18653/v1/2025.acl-long.528",
    pages = "10758--10774",
}

@misc{alpacaeval,
  title={Alpacaeval: An automatic evaluator of instruction-following models},
  author={Li, Xuechen and Zhang, Tianyi and Dubois, Yann and Taori, Rohan and Gulrajani, Ishaan and Guestrin, Carlos and Liang, Percy and Hashimoto, Tatsunori B},
  year={2023}
}

@misc{dsv32,
      title={DeepSeek-V3.2-Exp: Boosting Long-Context Efficiency with DeepSeek Sparse Attention}, 
      author={DeepSeek-AI},
      year={2025},
}

@article{llama3,
  title={The llama 3 herd of models},
  author={Dubey, Abhimanyu and Jauhri, Abhinav and Pandey, Abhinav and Kadian, Abhishek and Al-Dahle, Ahmad and Letman, Aiesha and Mathur, Akhil and Schelten, Alan and Yang, Amy and Fan, Angela and others},
  journal={arXiv e-prints},
  pages={arXiv--2407},
  year={2024}
}

@inproceedings{aqua,
    title = "Program Induction by Rationale Generation: Learning to Solve and Explain Algebraic Word Problems",
    author = "Ling, Wang  and
      Yogatama, Dani  and
      Dyer, Chris  and
      Blunsom, Phil",
    editor = "Barzilay, Regina  and
      Kan, Min-Yen",
    booktitle = "Proceedings of the 55th Annual Meeting of the Association for Computational Linguistics (Volume 1: Long Papers)",
    month = jul,
    year = "2017",
    address = "Vancouver, Canada",
    publisher = "Association for Computational Linguistics",
    url = "https://aclanthology.org/P17-1015/",
    doi = "10.18653/v1/P17-1015",
    pages = "158--167",
}

@article{gsm8k,
  title={Training verifiers to solve math word problems},
  author={Cobbe, Karl and Kosaraju, Vineet and Bavarian, Mohammad and Chen, Mark and Jun, Heewoo and Kaiser, Lukasz and Plappert, Matthias and Tworek, Jerry and Hilton, Jacob and Nakano, Reiichiro and others},
  journal={arXiv preprint arXiv:2110.14168},
  year={2021}
}

@inproceedings{gpqa,
title={{GPQA}: A Graduate-Level Google-Proof Q\&A Benchmark},
author={David Rein and Betty Li Hou and Asa Cooper Stickland and Jackson Petty and Richard Yuanzhe Pang and Julien Dirani and Julian Michael and Samuel R. Bowman},
booktitle={First Conference on Language Modeling},
year={2024},
url={https://openreview.net/forum?id=Ti67584b98}
}

@inproceedings{math500,
title={Let's Verify Step by Step},
author={Hunter Lightman and Vineet Kosaraju and Yuri Burda and Harrison Edwards and Bowen Baker and Teddy Lee and Jan Leike and John Schulman and Ilya Sutskever and Karl Cobbe},
booktitle={The Twelfth International Conference on Learning Representations (ICLR 2024)},
year={2024},
url={https://openreview.net/forum?id=v8L0pN6EOi}
}

@inproceedings{math,
title={Measuring Mathematical Problem Solving With the {MATH} Dataset},
author={Dan Hendrycks and Collin Burns and Saurav Kadavath and Akul Arora and Steven Basart and Eric Tang and Dawn Song and Jacob Steinhardt},
booktitle={Thirty-fifth Conference on Neural Information Processing Systems Datasets and Benchmarks Track (Round 2)},
year={2021},
url={https://openreview.net/forum?id=7Bywt2mQsCe}
}

@inproceedings{eta-sampling,
    title = "Truncation Sampling as Language Model Desmoothing",
    author = "Hewitt, John  and
      Manning, Christopher  and
      Liang, Percy",
    editor = "Goldberg, Yoav  and
      Kozareva, Zornitsa  and
      Zhang, Yue",
    booktitle = "Findings of the Association for Computational Linguistics: EMNLP 2022",
    month = dec,
    year = "2022",
    address = "Abu Dhabi, United Arab Emirates",
    publisher = "Association for Computational Linguistics",
    url = "https://aclanthology.org/2022.findings-emnlp.249/",
    doi = "10.18653/v1/2022.findings-emnlp.249",
    pages = "3414--3427",
}

@inproceedings{mirostat,
title={Mirostat: A neural text decoding algorithm that directly controls perplexity},
author={Sourya Basu and Govardana Sachitanandam Ramachandran and Nitish Shirish Keskar and Lav R. Varshney},
booktitle={International Conference on Learning Representations (ICLR 2021)},
year={2021},
url={https://openreview.net/forum?id=W1G1JZEIy5_}
}

@article{siml,
  title={Is there an optimal temperature and  top-p for code generation with paid LLM APIs?},
  author={Jan Siml},
  journal={Accessed: 2024-03-17},
  year={2024},
  url={https://siml.earth/scratchpad/llm_code_generation_experiment/}
}

@misc{qwen3,
      title={Qwen3 Technical Report}, 
      author={{Qwen Team}},
      year={2025},
      eprint={2505.09388},
      archivePrefix={arXiv},
      primaryClass={cs.CL},
      url={https://arxiv.org/abs/2505.09388}, 
}

@article{ackley1985,
  title={A learning algorithm for Boltzmann machines},
  author={Ackley, David H and Hinton, Geoffrey E and Sejnowski, Terrence J},
  journal={Cognitive science},
  volume={9},
  number={1},
  pages={147--169},
  year={1985},
  publisher={Elsevier}
}

@inproceedings{chen-ding-2023,
    title = "Probing the ``Creativity'' of Large Language Models: Can models produce divergent semantic association?",
    author = "Chen, Honghua  and
      Ding, Nai",
    editor = "Bouamor, Houda  and
      Pino, Juan  and
      Bali, Kalika",
    booktitle = "Findings of the Association for Computational Linguistics: EMNLP 2023",
    month = dec,
    year = "2023",
    address = "Singapore",
    publisher = "Association for Computational Linguistics",
    url = "https://aclanthology.org/2023.findings-emnlp.858/",
    doi = "10.18653/v1/2023.findings-emnlp.858",
    pages = "12881--12888"
}

@article{bellemare2024,
  title={Divergent creativity in humans and large language models},
  author={Bellemare-Pepin, Antoine and Lespinasse, Fran{\c{c}}ois and Th{\"o}lke, Philipp and Harel, Yann and Mathewson, Kory and Olson, Jay A and Bengio, Yoshua and Jerbi, Karim},
  journal={arXiv preprint arXiv:2405.13012},
  year={2024}
}

@article{ma2025,
  title={Estimating llm uncertainty with logits},
  author={Ma, Huan and Chen, Jingdong and Wang, Guangyu and Zhang, Changqing},
  journal={arXiv e-prints},
  pages={arXiv--2502},
  year={2025}
}

@article{real,
  author={Haw-Shiuan Chang and Nanyun Peng and Mohit Bansal and Anil Ramakrishna and Tagyoung Chung},
  title={REAL Sampling: Boosting Factuality and Diversity of Open-Ended Generation via Asymptotic Entropy},
  year={2024},
  journal={CarXiv  preprint arXiv:2406.07735.},
}

@inproceedings{acs2024,
    title = "Adaptive Contrastive Search: Uncertainty-Guided Decoding for Open-Ended Text Generation",
    author = "Garces Arias, Esteban  and
      Rodemann, Julian  and
      Li, Meimingwei  and
      Heumann, Christian  and
      A{\ss}enmacher, Matthias",
    editor = "Al-Onaizan, Yaser  and
      Bansal, Mohit  and
      Chen, Yun-Nung",
    booktitle = "Findings of the Association for Computational Linguistics: EMNLP 2024",
    month = nov,
    year = "2024",
    address = "Miami, Florida, USA",
    publisher = "Association for Computational Linguistics",
    url = "https://aclanthology.org/2024.findings-emnlp.885/",
    doi = "10.18653/v1/2024.findings-emnlp.885",
    pages = "15060--15080",
}

@inproceedings{guard2025,
    title = "{GUARD}: Glocal Uncertainty-Aware Robust Decoding for Effective and Efficient Open-Ended Text Generation",
    author = "Ding, Yuanhao  and
      Garces Arias, Esteban  and
      Li, Meimingwei  and
      Rodemann, Julian  and
      A{\ss}enmacher, Matthias  and
      Chen, Danlu  and
      Fan, Gaojuan  and
      Heumann, Christian  and
      Zhang, Chongsheng",
    editor = "Christodoulopoulos, Christos  and
      Chakraborty, Tanmoy  and
      Rose, Carolyn  and
      Peng, Violet",
    booktitle = "Findings of the Association for Computational Linguistics: EMNLP 2025",
    month = nov,
    year = "2025",
    address = "Suzhou, China",
    publisher = "Association for Computational Linguistics",
    url = "https://aclanthology.org/2025.findings-emnlp.380/",
    doi = "10.18653/v1/2025.findings-emnlp.380",
    pages = "7202--7226",
    ISBN = "979-8-89176-335-7",
}

@inproceedings{zhou-etal-2025-balancing,
    title = "Balancing Diversity and Risk in {LLM} Sampling: How to Select Your Method and Parameter for Open-Ended Text Generation",
    author = "Zhou, Yuxuan  and
      Keuper, Margret  and
      Fritz, Mario",
    editor = "Che, Wanxiang  and
      Nabende, Joyce  and
      Shutova, Ekaterina  and
      Pilehvar, Mohammad Taher",
    booktitle = "Proceedings of the 63rd Annual Meeting of the Association for Computational Linguistics (Volume 1: Long Papers)",
    month = jul,
    year = "2025",
    address = "Vienna, Austria",
    publisher = "Association for Computational Linguistics",
    url = "https://aclanthology.org/2025.acl-long.1278/",
    doi = "10.18653/v1/2025.acl-long.1278",
    pages = "26352--26365",
    ISBN = "979-8-89176-251-0",
}

@inproceedings{g2,
    title = "G2: Guided Generation for Enhanced Output Diversity in {LLM}s",
    author = "Ruan, Zhiwen  and
      Li, Yixia  and
      Liu, Yefeng  and
      Chen, Yun  and
      Luo, Weihua  and
      Li, Peng  and
      Liu, Yang  and
      Chen, Guanhua",
    editor = "Christodoulopoulos, Christos  and
      Chakraborty, Tanmoy  and
      Rose, Carolyn  and
      Peng, Violet",
    booktitle = "Proceedings of the 2025 Conference on Empirical Methods in Natural Language Processing",
    month = nov,
    year = "2025",
    address = "Suzhou, China",
    publisher = "Association for Computational Linguistics",
    url = "https://aclanthology.org/2025.emnlp-main.713/",
    doi = "10.18653/v1/2025.emnlp-main.713",
    pages = "14127--14145",
    ISBN = "979-8-89176-332-6",
}

@inproceedings{garces-arias-etal-2025-decoding,
    title = "Decoding Decoded: Understanding Hyperparameter Effects in Open-Ended Text Generation",
    author = "Garces Arias, Esteban  and
      Li, Meimingwei  and
      Heumann, Christian  and
      Assenmacher, Matthias",
    editor = "Rambow, Owen  and
      Wanner, Leo  and
      Apidianaki, Marianna  and
      Al-Khalifa, Hend  and
      Eugenio, Barbara Di  and
      Schockaert, Steven",
    booktitle = "Proceedings of the 31st International Conference on Computational Linguistics",
    month = jan,
    year = "2025",
    address = "Abu Dhabi, UAE",
    publisher = "Association for Computational Linguistics",
    url = "https://aclanthology.org/2025.coling-main.668/",
    pages = "9992--10020"
}

@inproceedings{wikitext1314,
title={Pointer Sentinel Mixture Models},
author={Stephen Merity and Caiming Xiong and James Bradbury and Richard Socher},
booktitle={International Conference on Learning Representations},
year={2017},
url={https://openreview.net/forum?id=Byj72udxe}
}

@inproceedings{
zhu2024improving,
title={Improving Open-Ended Text Generation via Adaptive Decoding},
author={Wenhong Zhu and Hongkun Hao and Zhiwei He and Yiming Ai and Rui Wang},
booktitle={Forty-first International Conference on Machine Learning (ICML 2024)},
year={2024},
}

@misc{deepseekv32,
      title={DeepSeek-V3.2: Pushing the Frontier of Open Large Language Models}, 
      author={DeepSeek-AI},
      year={2025},
}

@article{typical,
    title = "Locally Typical Sampling",
    author = "Meister, Clara  and
      Pimentel, Tiago  and
      Wiher, Gian  and
      Cotterell, Ryan",
    journal = "Transactions of the Association for Computational Linguistics",
    volume = "11",
    year = "2023",
    address = "Cambridge, MA",
    publisher = "MIT Press",
    url = "https://aclanthology.org/2023.tacl-1.7/",
    doi = "10.1162/tacl_a_00536",
    pages = "102--121"
}

\clearpage


\appendix

\section*{Appendix}

\section{Token Details}
\label{app:tokendetail}

\begin{figure}[htbp!]
\centering
\includegraphics[width=\linewidth]{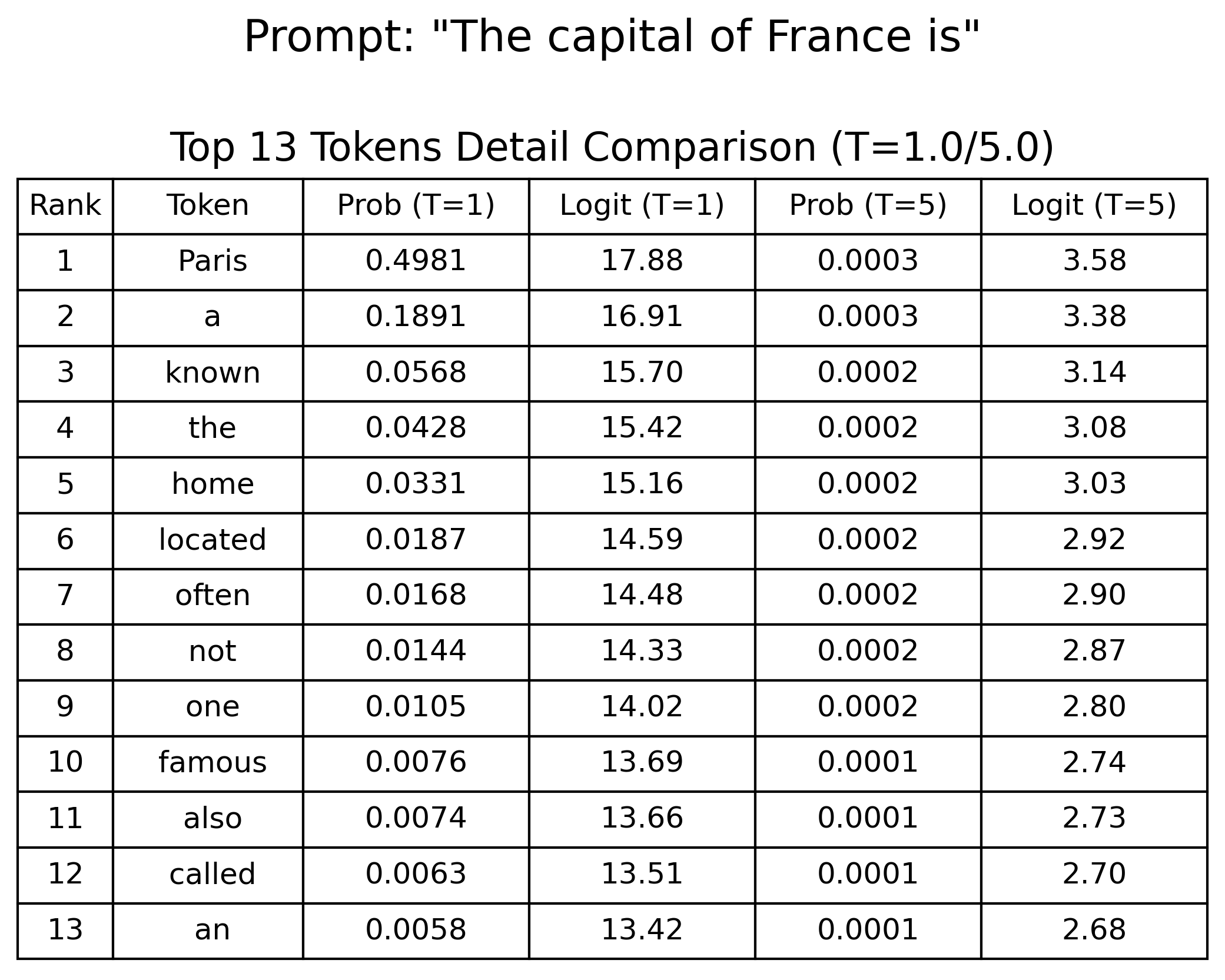}
\caption{Detailed probabilities and logits of the model's top-13 candidate tokens under the high-confidence prompt, at temperatures $T=1.0$ and $T=5.0$.}
\label{fig:htoken}
\end{figure}

In Figure~\ref{fig:one}, for the high-confidence prompt "The capital of France is", the model's top-13 candidate tokens are shown in Figure~\ref{fig:htoken} (only the top 13 are displayed due to space constraints and the presence of noise tokens beyond this point). As illustrated in Figure~\ref{fig:one}, the Min-$k$ method consistently selects the correct answer---"Paris"---under both temperature settings ($T=1.0$ and $T=5.0$), whereas other methods introduce varying degrees of noise tokens.

\begin{figure}[htbp!]
\centering
\includegraphics[width=\linewidth]{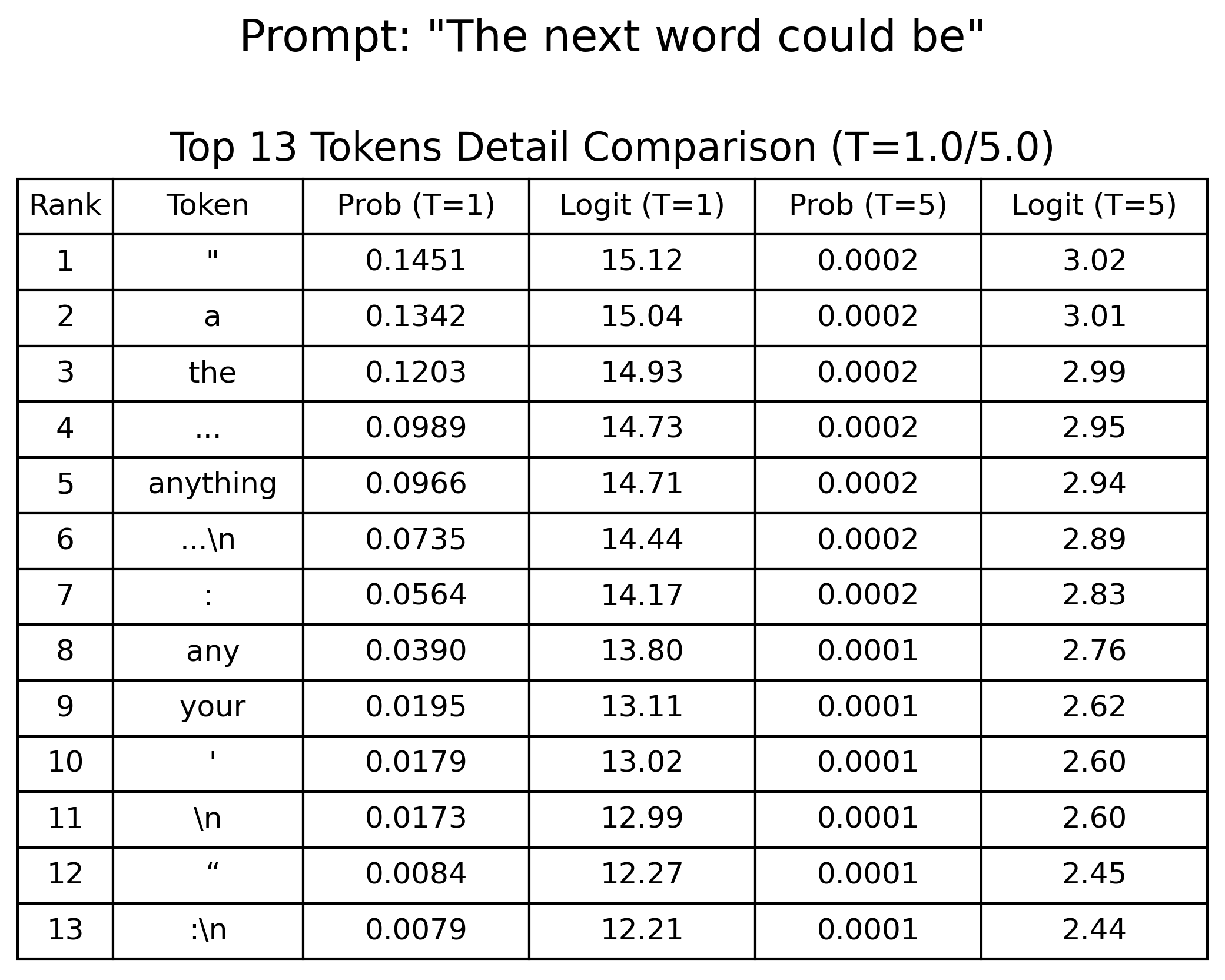}
\caption{Detailed probabilities and logits of the model's top-13 candidate tokens under the low-confidence prompt, at temperatures $T=1.0$ and $T=5.0$.}
\label{fig:ltoken}
\end{figure}

For the low-confidence prompt "The next word could be" in Figure~\ref{fig:one}, the model's top-13 candidate tokens are shown in Figure~\ref{fig:ltoken}. As illustrated in Figure~\ref{fig:one}, the Min-$k$ method selects 8 tokens. Although some of these candidates exhibit weak semantic relevance, they still lie within the model's high-confidence region. In contrast, other methods place their truncation thresholds farther out, thereby including a larger number of low-confidence tokens with weak semantic coherence. This demonstrates that Min-$k$ effectively identifies the pronounced drop in semantic confidence and selects an adaptive truncation point that suppresses semantically irrelevant noise tokens—prioritizing the overall quality of the candidate set over the perfection of individual tokens.

\section{Visualization and Dynamic Behavior}
\label{sec:visual}
\begin{figure}[!ht]
\centering
\includegraphics[width=\linewidth]{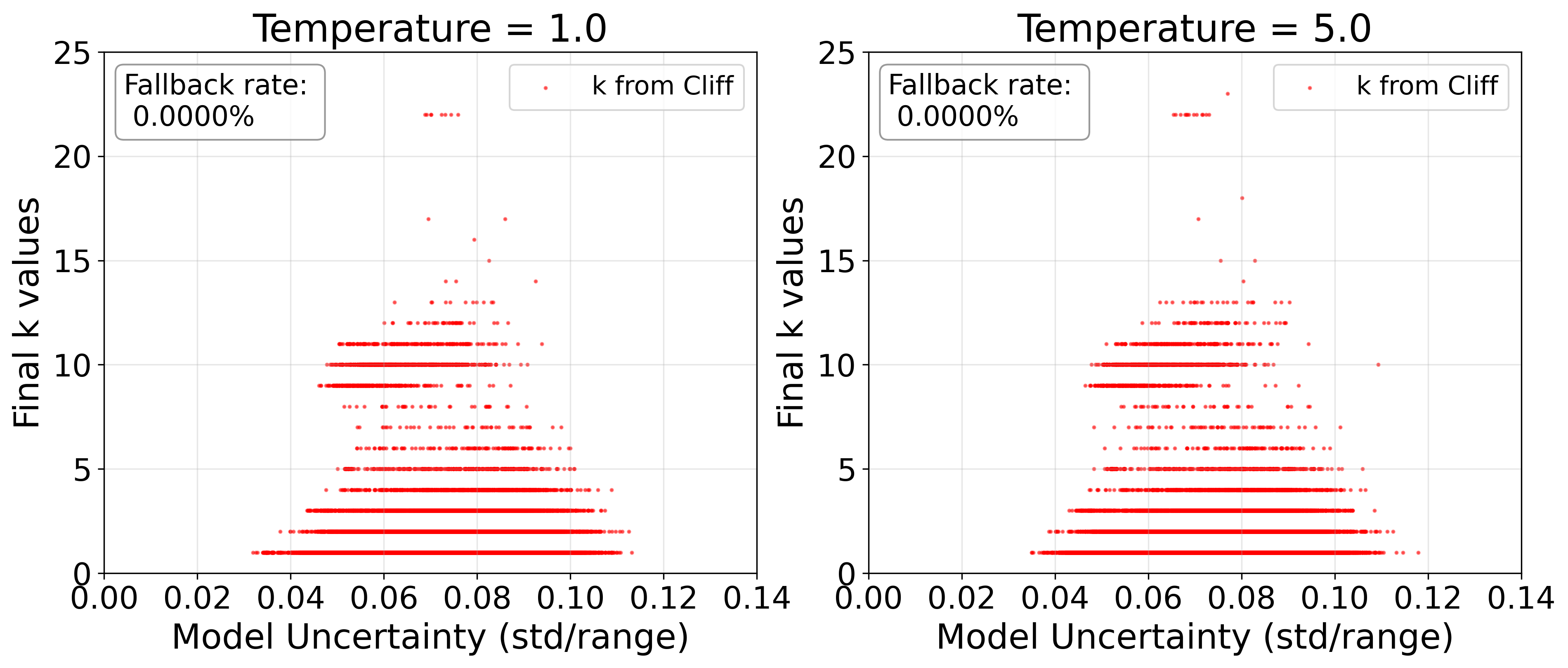}
\caption{Visualization of candidate set size $k$ across decoding steps under different temperature settings. Left: $T=1.0$. Right: $T=5.0$. The distributions are nearly identical, indicating strong temperature invariance.}
\label{fig:distribution}
\end{figure}

To empirically validate the dominance of $k_{\text{cliff}}$ over $k_{\text{fallback}}$ and observe dynamic adaptation, we conducted a study using 500 random samples from the Wikitext-103 \cite{wikitext1314}, with 256 generated tokens each, setting $\tau=30$ to highlight the fallback. Figure~\ref{fig:distribution} shows the distribution of $k$ values during generation under temperature $T=1.0$ and $T=5.0$. The results demonstrate that the final decisions are primarily governed by $k_{\text{cliff}}$; the fallback mechanism is rarely triggered except in extremely flat distributions. Moreover, the $k$ values adapt dynamically from 1 to 20+ and remain almost identical across different temperatures, confirming the strong temperature invariance of our approach.

\section{Dataset Processing}
\label{a:datasetprocessing}

\subsection{Reasoning Datasets}
\label{a:reasoning}
We follow the datasets adopted in \citet{topn}, but our data preprocessing procedures \textbf{differ slightly}. Each dataset is handled according to its specific structural characteristics. The common principle is to formulate every problem as an open-ended generation task, from which the model’s predicted answer is extracted and compared with the ground-truth label to compute accuracy. The code for the following four evaluation methods has been publicly released.

\subsection*{AQuA Dataset} The AQuA dataset focuses on algebraic word problems, each accompanied by five multiple-choice options. It primarily evaluates a model’s ability to comprehend and make decisions in a multiple-choice setting. For each instance, we structure the question and its five options (A--E) within the prompt, and introduce a clear output instruction: \textit{\textbf{Your response must end with "The answer is (X)" where X is one of A, B, C, D, or E.}} This instruction enforces the model to provide a definitive and unique choice after its reasoning process, avoiding vague or uncertain responses. To parse the model’s final choice, we employ a flexible regular expression:
\begin{verbatim}
The answer is\s*\(?([A-E])\)?
\end{verbatim}

This pattern not only matches the guiding phrase but also reliably captures the target letter representing the model’s selected option. The extracted prediction is then compared against the gold answer through exact string matching to determine correctness. This design improves parsing robustness and ensures accurate evaluation.

\subsection*{GSM8K Dataset}
The GSM8K dataset centers on elementary-level mathematical word problems, emphasizing multi-step arithmetic reasoning. To ensure consistent and automatable evaluation, we use careful prompt engineering to constrain the model’s output. Each prompt concludes with a mandatory instruction: \textit{\textbf{Your response must end with "The final answer is (answer)".}} This directive guides the model to clearly separate its reasoning process from the final numerical answer. We extract the final answer using the following regular expression: 
\begin{verbatim}
The final answer is\s*([-]?\d{1,3}(?:,
\d{3})*\.?\d*)
\end{verbatim}

This pattern accurately captures numerical outputs, including negative numbers, thousand separators, and decimals. The ground-truth answer is taken from the dataset’s field marked with $\#\#\#\#$. Correctness is determined by comparing the extracted prediction with the reference value.

\subsection*{GPQA-main Dataset}
The GPQA-main dataset consists of highly specialized and challenging questions in biology, physics, and chemistry, designed to test a model’s ability to solve complex problems. We convert the columns Correct Answer and Incorrect Answer 1/2/3 into a multiple-choice format. In our implementation, we randomly shuffle the four answer options to ensure that the position of the correct answer is unpredictable, thereby mitigating the model's positional bias toward specific choices. This design enforces the model to rely on its inherent knowledge and reasoning capabilities to make accurate selections. We add a strict instruction to the prompt: \textit{\textbf{Your response *must* end with "The final answer is (answer)". For example: (Question and your reasoning) The final answer is (A).}} Example prompts are also provided to reduce deviation in model behavior. For answer extraction, we use the regex:
\begin{verbatim}
The final answer is\s*\(([A-D])\)\.?\s*$
\end{verbatim}

This mechanism not only enforces that option identifiers are enclosed within parentheses but also employs the \texttt{\.?\textbackslash s*\$} assertion to guarantee that the response adheres to a strictly defined ending format. Consequently, this allows for a fairer and more precise evaluation of the model's genuine capabilities in expert-level question answering tasks. Accuracy is computed through exact matching between the predicted and ground-truth options.

\subsection*{MATH500 Dataset}
The MATH500 dataset focuses on complex mathematical competition problems, which often require advanced reasoning and produce diverse answer formats, including LaTeX fractions, radicals, and symbolic expressions. To ensure reliable evaluation, we implement a comprehensive answer normalization pipeline. All LaTeX-specific commands (e.g., \texttt{\textbackslash frac}, \texttt{\textbackslash left}, \texttt{\textbackslash right}) are removed, symbolic formats are converted to standard text forms (e.g., \texttt{\textbackslash frac} $\rightarrow$ \texttt{/}), excessive spaces are stripped, and the entire text is lowercased. This procedure maps mathematically equivalent expressions to a unique canonical representation. The prompt enforces the same instruction as before: \textit{\textbf{Your response must end with "The final answer is (answer)".}} After generation, we extract the answer using the following regular expression:
\begin{verbatim}
The final answer is[:\s]+(.+?)(?:\.|$)
\end{verbatim}

We then apply a two-stage matching strategy: Exact string comparison between normalized predictions and references. If the strings do not match, both are parsed as floating-point numbers, and correctness is determined by checking whether they are equal within a very small tolerance. This dual validation mechanism ensures both robustness and fairness when evaluating complex mathematical answers.

\subsection{Creative Writing}
\label{a:creative}

We adopt the \citet{topn} and \citet{minp} creative writing evaluation paradigms to assess the generative quality of our model on open-ended, creative writing tasks. To ensure a more comprehensive and fair evaluation, we employ an advanced large language model (\textbf{DeepSeek-V3.2-Exp}) as the judge. 
Unlike traditional word-overlap-based metrics (e.g., BLEU, ROUGE), our method focuses on qualitative aspects of text generation, such as creativity, depth, relevance, and instruction adherence. Our evaluation follows a core A/B comparison framework. Each decoding strategy under evaluation generates a "candidate" response, which is then pairwise-compared with a unified, high-quality "reference" response.

Specifically, we adopt 500 instruction prompts from the \texttt{AlpacaEval2.0} \cite{alpacaeval} dataset as creative-writing inputs, following \citet{topn} and \citet{minp}. For each prompt, we generate a reference answer using greedy decoding, which serves as the baseline. We then evaluate various stochastic decoding strategies by generating candidate responses under different temperature settings ($T = 1.0, 2.0, 5.0$ and $10.0$).

The core of the evaluation process is conducted by the DeepSeek-V3.2-Exp model acting as an impartial judge. Each comparison includes the instruction, the reference answer (A), and the candidate answer (B). To eliminate position bias, the presentation order of A and B is randomized. The judging LLM is provided with detailed evaluation instructions, guiding it to compare both responses holistically across multiple dimensions—\emph{helpfulness}, \emph{relevance}, \emph{accuracy}, \emph{creativity}, and \emph{level of detail}—and to determine which response is superior.

The evaluation metric is the \textbf{Win Rate}, defined as the percentage of test prompts where a candidate response generated by a specific decoding strategy is preferred over the greedy baseline by the judging LLM. This metric not only quantifies the relative improvement of each decoding algorithm but also provides insights into their effectiveness in eliciting creativity and generating richer, more profound content.

\begin{table*}[!htbp]
\centering
\resizebox{\textwidth}{!}{\begin{tabular}{ccc}
\toprule
\multicolumn{3}{c}{Analysis of Component Contributions} \\ 

\midrule
\rowcolor{Best}Full-Min-$k$ & 76.15 & \textit{Complete methodical, and balances all factors.} \\
w/o Weight (1/$i$) & 64.14 & \textit{Validate the necessity of Head-bias. Removing it increases $k$ and introduces more noise.} \\
w/o Range Norm ($R_l$) & 41.02 & \textit{Validate Temperature Invariance. Performance drops significantly at high $T$ when removed.} \\
w/o Fallback & 76.15 & \textit{Validate Robustness. Prevents degradation under extremely flat distributions caused by $k=1$.}  \\
\bottomrule
\end{tabular}
}
\caption{\textbf{Ablation study of Min-$k$ design components on the GSM8K dataset ($T=4.0$).} We compare the proposed Full Min-$k$ against variants without position weighting ($1/i$), range normalization ($R_l$), and the fallback mechanism. The results demonstrate that range normalization is fundamental for temperature invariance (as seen in the collapse of the "w/o Range Norm" variant), while position weighting effectively filters tail noise.}
\label{tab:ablation}
\end{table*}

\begin{table*}[t]
\centering 
\resizebox{0.65\textwidth}{!}{\begin{tabular}{lccccc}
    \toprule
    \multirow{2}{*}{\textbf{Decay Function}} & \multicolumn{2}{c}{\textbf{GSM8K Accuracy (\%)}} & & \multicolumn{2}{c}{\textbf{Creative Writing Win Rate (\%)}} \\
    \cmidrule{2-3} \cmidrule{5-6}
    & $T=1.0$ & $T=4.0$ & & $T=1.0$ & $T=4.0$ \\
    \midrule
    $1/i^0$ (No Weighting) & 75.97 & 64.14 & & \underline{52.80} & 11.00 \\
    $1/i^{0.5}$ (Sqrt)     & 77.41 & 71.57 & & 49.00 & 49.20 \\
    \rowcolor{Best}\textbf{$1/i^1$ (Linear, Ours)} & \underline{77.39} & \textbf{76.15} & & \textbf{58.60} & \textbf{53.20} \\
    $1/i^2$ (Square)       & \textbf{78.54} & \underline{76.04} & & 50.20 & \underline{50.40} \\
    $\log(i)$              & 74.07 & 72.71 & & 51.60 & 48.20 \\
    \bottomrule
\end{tabular}}
\caption{\textbf{Impact of different decay functions on reasoning (GSM8K) and creative writing (AlpacaEval) tasks.} While aggressive decay ($1/i^2$) slightly favors strict reasoning by filtering tail noise, it harms generation quality. The proposed linear decay ($1/i$) achieves the \textbf{best overall generalization} across domains and temperatures.}
\label{tab:decay_ablation}
\end{table*}

\section{Ablation Study}
\label{app:ablation}

\subsection{Component Analysis}
\label{app:component}

To rigorously evaluate the contribution of each design choice within the Min-k framework, we conduct an ablation study on the GSM8K dataset under a high-temperature setting ($T = 4.0$), with results shown in Table~\ref{tab:ablation}. The results provide strong evidence for the necessity of our geometric constraints. Specifically, removing the \textbf{positional weighting decay} ($1/i$) leads to a sharp performance drop from 74.53\% to 64.14\%, confirming that semantic boundaries predominantly reside near the head of the distribution and that penalizing tail fluctuations is critical for noise filtering. Additionally, \textbf{dynamic range normalization} ($R_l$) is indispensable for achieving temperature invariance; its absence causes catastrophic degradation (down to 41.02\%) because the raw logit gradients vanish under temperature scaling, rendering "cliffs" undetectable. Omitting the \textbf{fallback mechanism} yields identical performance (74.53\%), indicating that the core cliff-detection mechanism is already sufficiently robust for structured reasoning tasks and can operate independently, while the fallback serves as a theoretical safety guardrail that prevents degradation without interfering with correct reasoning paths.

\noindent\textbf{Rationale for the default $\tau = 3.0$:} While our sensitivity analysis (Sec~\ref{sec:visual} and \ref{subsec:analysis}) shows stable performance across a wide range of $\tau$ values (1.0--10.0), we select 3.0 as the default. Intuitively, a candidate set size of $k < 3$ approaches deterministic behavior, limiting creativity. Setting $\tau \approx 3$ guarantees a minimal degree of freedom for the sampling without requiring complex tuning.

\subsection{Sensitivity to Decay Functions}
\label{app:decay}
We investigated the impact of the weighting function form on model performance, comparing power-law decays ($w_i \propto 1/i^\alpha$ with varying $\alpha$) and a logarithmic alternative ($w_i \propto 1/\log(i)$). Table~\ref{tab:decay_ablation} presents the results across reasoning (GSM8K) and open-ended generation (Creative Writing) tasks.

The results reveal an intriguing trade-off between precision and diversity governed by the decay speed. 
\textbf{Aggressive decay} (e.g., $1/i^2$) yields the highest accuracy on GSM8K by rigorously suppressing long-tail noise, which is beneficial for strict reasoning. However, this comes at the cost of creativity: $1/i^2$ underperforms on the Creative Writing task, likely because it over-truncates plausible but lower-ranked tokens that contribute to linguistic diversity. 
Conversely, \textbf{weak decay} functions—including $1/i^0$, $1/i^{0.5}$, and $1/\log(i)$—fail to sufficiently filter noise, particularly at high temperatures. For instance, the unweighted variant ($1/i^0$) collapses to an 11.00\% win rate at $T=4.0$, and $\log(i)$ consistently lags behind linear.

\textbf{Our proposed linear decay ($1/i$) emerges as the most robust strategy.} It achieves the highest win rates in Creative Writing (up to 58.60\%) while maintaining competitive reasoning accuracy (within 1.5\% of the aggressive optimum). This confirms that $1/i$ strikes the optimal balance for general-purpose LLM decoding.

\section{Semantic Noise Evaluation Protocol}
\label{app:noise_eval}

To rigorously quantify the ``semantic collapse'' phenomenon observed in high-temperature generation, we designed an LLM-based evaluation pipeline.

\paragraph{Definition of Semantic Noise.} We define the ``Semantic Collapse Point'' as the index of the first sentence where the generated text transitions from logical reasoning into degeneracy. Degeneracy explicitly includes:

\begin{itemize}[noitemsep]
    \item \textbf{Nonsense/Gibberish:} Sequences of unrelated words, garbled text, or non-linguistic symbols.
    \item \textbf{Infinite Repetition:} Repeating the same phrase, sentence, or reasoning step in an infinite loop.
    \item \textbf{Irrelevant Hallucination:} Discussing topics or contexts completely unrelated to the given mathematical problem.
\end{itemize}

\paragraph{Metric Calculation.} For a generated response $R$ consisting of a sequence of sentences $S = [s_1, s_2, \dots, s_n]$, let $k$ be the index of the collapse point identified by the judge. The \textbf{Semantic Noise Rate (SNR)} is calculated as the length ratio of the collapsed text segment to the total text length:
\begin{equation}
    \text{SNR}(R) = 
    \begin{cases} 
    0 & \text{if } R \text{ is coherent,} \\
    \frac{\sum_{i=k}^{n} \text{len}(s_i)}{\sum_{j=1}^{n} \text{len}(s_j)} & \text{if collapse starts at } k.
    \end{cases}
\end{equation}

\paragraph{LLM-as-a-Judge Implementation.} We utilized the DeepSeek-V3.2-Exp model as the evaluator due to its strong instruction-following and reasoning capabilities. The model was prompted to analyze the generated text sentence-by-sentence and output the exact index of collapse in a structured JSON format. We processed 128 samples from the GSM8K dataset for each method-temperature pair (5 methods $\times$ 5 temperatures = 25 configurations), ensuring statistical significance. The core instruction prompt used for evaluation is as follows:
\textit{\textbf{"You are an expert evaluator of mathematical reasoning quality. Your task is to identify the exact `Semantic Collapse Point' in a model-generated solution. It occurs when the text transitions from logical reasoning into nonsense, infinite repetitive loops, or irrelevant topics. You must output the index of the FIRST sentence where the collapse begins, or -1 if the entire solution is coherent."}}

\section{Human Evaluation Instructions}
\label{app:F}

\begin{tcolorbox}[
    colback=gray!10,     
    colframe=black!70,   
    arc=8pt,             
    boxrule=1pt,         
    title=Evaluation Instructions,  
    fonttitle=\bfseries, 
]
Thank you for participating in this evaluation!

\subsubsection*{Task Overview}
You will compare pairs of AI-generated responses and indicate which one is better. Each row contains a prompt and two responses (A and B) from different generation methods.

\subsubsection*{Evaluation Criteria}
Consider the following factors when making your judgment:
\begin{enumerate}
    \item \textbf{Relevance} -- Does the response address the prompt appropriately?
    \item \textbf{Coherence} -- Is the response well-organized and logically structured?
    \item \textbf{Completeness} -- Does the response fully answer the question?
    \item \textbf{Accuracy} -- Is the information provided correct?
    \item \textbf{Quality} -- Is the writing clear, fluent, and human-like?
\end{enumerate}

\subsubsection*{How to Respond}
In the `Your Preference' column, enter:
\begin{itemize}
    \item \textbf{A} -- if Response A is clearly better
    \item \textbf{B} -- if Response B is clearly better
    \item \textbf{Tie} -- if both responses are roughly equal in quality
\end{itemize}

\subsubsection*{Important Notes}
\begin{itemize}
    \item Read both responses completely before deciding
    \item Focus on overall quality, not just length
    \item Ideally, use `Tie' only when you genuinely cannot distinguish quality
\end{itemize}
\end{tcolorbox}

\begin{table*}[t!]
\centering
\resizebox{1\textwidth}{!}{\begin{tabular}{clcccccccccccc}
\toprule
\multirow{2}{*}{Dataset} & Model & \multicolumn{6}{c}{LLaMA3-70B-Instruct} & \multicolumn{6}{c}{Qwen3-30B-Instruct}  \\
\cmidrule(lr){3-8} \cmidrule(lr){9-14}
 & $T$ & 1.0 & 2.0 & 3.0 & 4.0 & 5.0 & 10.0 &1.0 & 2.0 & 3.0 & 4.0 & 5.0 & 10.0  \\
\midrule
\multirow{5}{*}{AQuA} 
 & Top-$k$ & 73.23 & 70.87 & 54.33 & 27.56 & 15.75 & 3.54 & 54.55 & 50.26 & 42.62 & 19.23 & 4.23 & 1.24 \\
 & Top-$p$ & 74.02 & 68.90 & 35.83 & 10.24 & 4.33 & 0.00 & 57.02 & 55.26 & 37.95 & 5.83 & 1.79 & 0.00 \\
 & Min-$p$ & 74.41 & 70.87 & 65.35 & 42.52 & 16.54 & 0.79 & 59.67 & 56.84 & 49.57 & 29.63 & 10.35 & 0.68 \\
 & Top-$n\sigma$ & \cellcolor{Best}{\textbf{74.80}} & \cellcolor{Best}{\textbf{74.80}} & 70.47 & 71.65 & \cellcolor{Best}{\textbf{72.83}} & 72.83 & \cellcolor{Best}{\textbf{61.26}} & \cellcolor{Best}{\textbf{60.83}} & 59.87 & 59.23 & 60.22 & \cellcolor{Best}{\textbf{60.12}} \\ 
 & Min-$k$ & 73.23 & 73.62 & \cellcolor{Best}{\textbf{72.83}} & \cellcolor{Best}{\textbf{72.83}} & 72.05 & \cellcolor{Best}{\textbf{73.62}} & 59.46 & 59.06 & \cellcolor{Best}{\textbf{60.24}} & \cellcolor{Best}{\textbf{60.63}} & \cellcolor{Best}{\textbf{60.79}} & 59.84 \\
 \hline
 \multirow{5}{*}{GPQA-main} 
 & Top-$k$ & \cellcolor{Best}{\textbf{39.73}} & 38.17 & 34.82 & 20.76 & 8.26 & 0.22 & \cellcolor{Best}{\textbf{26.56}} & 21.87 & 14.73 & 7.81 & 3.34 & 0.00 \\ 
& Top-$p$ & 35.71 & 37.50 & 30.13 & 8.93 & 3.35 & 0.00 & 22.54 & 21.20 & 11.16 & 3.34 & 0.00 & 0.00 \\
& Min-$p$ & \cellcolor{Best}{\textbf{39.73}} & 37.95 & 39.06 & 29.01 & 17.19 & 0.00 & 22.55 & 22.43 & 20.11 & 14.88 & 8.56 & 0.51 \\
& Top-$n\sigma$ & 37.72 & \cellcolor{Best}{\textbf{38.39}} & 38.62 & \cellcolor{Best}{\textbf{39.95}} & 38.84 & 38.39 & 23.66 & 23.88 & 23.43 & 23.21 & 23.43 & 23.88 \\
& Min-$k$ & 38.62 & 37.95 & \cellcolor{Best}{\textbf{39.96}} & 39.39 & \cellcolor{Best}{\textbf{40.85}} & \cellcolor{Best}{\textbf{40.85}} & 25.00 & \cellcolor{Best}{\textbf{25.22}} & \cellcolor{Best}{\textbf{24.33}} & \cellcolor{Best}{\textbf{24.33}} & \cellcolor{Best}{\textbf{24.55}} & \cellcolor{Best}{\textbf{24.33}} \\
 \hline
\multirow{5}{*}{GSM8K} 
& Top-$k$ & 92.64 & 91.13 & 81.20 & 36.92 & 8.57 & 0.00 & 95.83 & 95.15 & 94.62 & 89.46 & 62.02 & 0.00 \\ 
& Top-$p$ & \cellcolor{Best}{\textbf{92.95}} & 92.04 & 64.22 & 9.63 & 0.83 & 0.00 & 95.60 & 95.00 & 92.95 & 77.56 & 52.34 & 0.00 \\
& Min-$p$ & 92.87 & 91.66 & 87.95 & 58.98 & 20.77 & 0.00 & 95.60 & 95.00 & 94.92 & 92.57 & 84.69 & 0.45 \\
& Top-$n\sigma$ & 92.87 & \cellcolor{Best}{\textbf{93.25}} & \cellcolor{Best}{\textbf{94.01}} & \cellcolor{Best}{\textbf{92.79}} & \cellcolor{Best}{\textbf{92.49}} & \cellcolor{Best}{\textbf{92.95}} & \cellcolor{Best}{\textbf{95.68}} & \cellcolor{Best}{\textbf{95.60}} & 95.37 & \cellcolor{Best}{\textbf{95.45}} & 95.30 & 94.69 \\
& Min-$k$ & \cellcolor{Best}{\textbf{92.95}} & 92.34 & 92.65 & 92.42 & 91.96 & 91.89 & 95.60 & 95.07 & \cellcolor{Best}{\textbf{95.60}} & \cellcolor{Best}{\textbf{95.45}} & \cellcolor{Best}{\textbf{95.37}} & \cellcolor{Best}{\textbf{95.37}} \\
 \hline
 \multirow{5}{*}{MATH500} 
 & Top-$k$ & 44.00 & 35.60 & 21.40 & 3.40 & 1.40 & 0.00 & \cellcolor{Best}{\textbf{61.20}} & 58.80 & 53.40 & 44.40 & 15.80 & 0.00 \\ 
 & Top-$p$ & 44.20 & 36.80 & 12.20 & 0.60 & 0.00 & 0.00 & 59.40 & 58.00 & 53.00 & 29.80 & 2.40 & 0.00 \\
 & Min-$p$ & 43.80 & 40.60 & 29.40 & 10.20 & 1.40 & 0.00 & 59.80 & 59.20 & 57.00 & 49.00 & 12.20 & 0.00 \\
 & Top-$n\sigma$ & 43.40 & 43.20 & 42.40 & 41.80 & \cellcolor{Best}{\textbf{43.40}} & 42.20 & 60.60 & 59.20 & 58.00 & 58.80 & 59.00 & 57.80 \\
 & Min-$k$ & \cellcolor{Best}{\textbf{46.40}} & \cellcolor{Best}{\textbf{43.60}} & \cellcolor{Best}{\textbf{43.00}} & \cellcolor{Best}{\textbf{43.20}} & 41.40 & \cellcolor{Best}{\textbf{46.40}} & \cellcolor{Best}{\textbf{61.20}} & \cellcolor{Best}{\textbf{60.00}} & \cellcolor{Best}{\textbf{61.00}} & \cellcolor{Best}{\textbf{60.00}} & \cellcolor{Best}{\textbf{59.80}} & \cellcolor{Best}{\textbf{59.40}} \\

\bottomrule
\end{tabular}}
\caption{Exact Match (\%) for different temperatures (1.0--10.0) and sampling strategies on AQuA, GPQA-main, GSM8K and MATH500 using the LLaMA-70B-Instruct and Qwen3-30B-Instruct. Best results in \textbf{bold}.}
\label{tab:largetable}
\end{table*}

\section{Additional model reasoning results}
\label{app:more model}

We also evaluated the inference results on LLaMA-70B-Instruct and Qwen3-30B-Instruct, as shown in Table \ref{tab:largetable}.

Table 6 presents a comprehensive empirical evaluation of exact match accuracy across diverse reasoning benchmarks, elucidating the critical impact of temperature scaling on various sampling strategies. A prominent observation is the exceptional resilience of the Min-$k$ method against the catastrophic degradation typically observed in conventional sampling techniques at high temperatures. While standard approaches such as Top-$k$, Top-$p$, and Min-$p$ suffer a precipitous decline in performance as temperature ($T$) increases—often plummeting to near-zero accuracy at $T=5.0$ and $T=10.0$ due to the flattening of the probability distribution and the intrusion of long-tail noise—Min-$k$ maintains robust performance levels remarkably consistent with, or even exceeding, those at lower temperature settings. For instance, on the AQuA dataset utilizing LLaMA-3-70B-Instruct, while Top-$p$ accuracy collapses from 74.02\% at $T=1.0$ to 0.00\% at $T=10.0$, Min-$k$ retains a high accuracy of 73.62\%, effectively neutralizing the adverse effects of excessive stochasticity while preserving the model's reasoning capabilities.

Furthermore, Min-$k$ demonstrates superior efficacy compared to the robust Top-$n\sigma$ baseline, establishing itself as the dominant strategy across both model architectures and all evaluated datasets. As evidenced by the concentration of bolded values in the Min-$k$ rows, this method consistently achieves state-of-the-art results, particularly in challenging scenarios requiring complex reasoning, such as GPQA-main. In this specific benchmark, under the extreme condition of $T=10.0$ for LLaMA-3-70B, Min-$k$ not only outperforms Top-$n\sigma$ (40.85\% vs. 38.39\%) but also surpasses the peak performance of Top-$k$ and Top-$p$ achieved at standard temperatures. This trend is mirrored in the Qwen3-30B-Instruct experiments, where Min-$k$ dominates the MATH500 benchmark at high temperatures ($59.40\%$ at $T=10.0$ versus $0.00\%$ for Top-$p$), validating Min-$k$ as a highly generalizable and effective truncation mechanism that enables models to leverage the generation quality of high temperatures without sacrificing semantic coherence or logical precision.

\paragraph{Universality Across Model Scales.} To investigate the generalizability of our findings, we extended the evaluation to smaller architectures, specifically LLaMA-3-8B-Instruct and Qwen3-4B-Instruct, as detailed in Table \ref{tab:table1}. A cross-examination of the results for both large-scale (Table 6) and smaller-scale models reveals that the efficacy of the Min-$k$ strategy is remarkably scale-invariant. Despite the reduced parameter count, which typically renders models more susceptible to incoherence under high-entropy conditions, Min-$k$ consistently prevents the catastrophic performance collapse observed in standard sampling methods. For instance, on the GSM8K benchmark, the LLaMA-3-8B model with Top-$p$ sampling degrades to $0.00\%$ accuracy at $T=10.0$, whereas Min-$k$ sustains a robust performance of $74.79\%$, effectively mirroring the stability seen in its 70B counterpart. Furthermore, comparisons against the Top-$n\sigma$ baseline on the Qwen3-4B model reinforce the superiority of our approach; on the AQuA dataset at $T=10.0$, Min-$k$ achieves $79.13\%$, surpassing both Top-$n\sigma$ ($77.17\%$) and the collapsed standard baselines. This consistent dominance across varying model capacities suggests that Min-$k$ functions as a fundamental correction to the sampling distribution tail, allowing even resource-constrained models to benefit from the accuracy of high temperatures without succumbing to noise, thereby establishing Min-$k$ as a universally robust truncation mechanism for large language models.

\end{document}